\documentclass[twocolumn]{article}

\usepackage{PRIMEarxiv}

\usepackage[utf8]{inputenc}
\usepackage[T1]{fontenc}
\usepackage{hyperref}
\usepackage{url}
\usepackage{booktabs}
\usepackage{amsfonts}
\usepackage{nicefrac}
\usepackage{microtype}
\usepackage{fancyhdr}
\usepackage{graphicx}
\usepackage{enumitem}
\usepackage{xcolor}
\usepackage{listings}
\graphicspath{{media/}}
\usepackage{lipsum}
\usepackage{amsmath}
\usepackage{xfrac}
\usepackage{subcaption}
\usepackage{makecell}
\usepackage{multirow}

\usepackage{caption}
\newcommand{\link}[2]{\href{#1}{#2}}

\title{MNet++: Extended 2D/3D Networks for Anisotropic Medical Image Segmentation \\
}

\author{
 Kirsten Odendaal \\
  School of Computing\\
   Georgia Institute of Technology\\
  Atlanta, GA, USA \\
  \texttt{kodendaal3@gatech.edu} \\
   \And
 Rade Bajic \\
  School of Computing\\
   Georgia Institute of Technology\\
  Atlanta, GA, USA \\
  \texttt{rbajic3@gatech.edu} \\
}

\begin{document}
\twocolumn[{
  \begin{@twocolumnfalse}

	\maketitle


\vspace{-2em}

\begin{abstract}

This work demonstrates a full reproduction and extension of MNet, a hybrid 2D/3D convolutional network designed for anisotropic medical image segmentation. The original architecture was re-implemented within the nnU-Net framework to verify its reported performance and robustness to variable voxel spacing known as anisotropy. Experiments were conducted on PROMISE (prostate MRI) and a controlled subset of LiTS (liver CT) under matched preprocessing and compute constraints. The reproduced MNet achieved a Dice similarity coefficient (DSC) of 89.0 $\pm$ 0.9 \% on PROMISE - within 0.8 \% of the published result and 94.3 $\pm$ 1.9 \% / 54.6 $\pm$ 3.1 \% for liver and tumor segmentation on LiTS, respectively. Two lightweight extensions were further introduced: (1) a learned Fusion Gating mechanism enabling adaptive 2D–3D feature blending, and (2) a VMamba state-space module for efficient long-range depth modelling. The Spatial Gating variant improved DSC by +0.8 \% with < 3 \% inference overhead, while VMamba improved performance consistency, reducing PROMISE Dice variation to ±0.7\% and achieving the strongest LiTS liver performance at 95.8\% Dice. Both extensions preserved MNet’s robustness to anisotropy ($\Delta$Dice $\approx$ 1.5 \% across 1–4 mm voxel spacing). Overall, the study confirms MNet’s reproducibility and demonstrates that adaptive fusion and state-space modelling have the potential to further strengthen segmentation reliability under anisotropic conditions. However, further tests are required to provide definitive conclusions.

The public github repository and video walk-through can be found here: \link{https://github.gatech.edu/kodendaal3/bd4h_mnet_b4.git}{\textbf{GitHub}} and \link{https://gtvault-my.sharepoint.com/:v:/g/personal/kodendaal3_gatech_edu/IQCqnuMKVMNFTZKzzQAv55LgAfGCsv7uiExv-MpQHGBaM2c?nav=eyJyZWZlcnJhbEluZm8iOnsicmVmZXJyYWxBcHAiOiJPbmVEcml2ZUZvckJ1c2luZXNzIiwicmVmZXJyYWxBcHBQbGF0Zm9ybSI6IldlYiIsInJlZmVycmFsTW9kZSI6InZpZXciLCJyZWZlcnJhbFZpZXciOiJNeUZpbGVzTGlua0NvcHkifX0&e=M8eFdO}{\textbf{Video}} URL.


\end{abstract}

  \end{@twocolumnfalse}
  \vspace{1.5em}
}]



\section{Introduction}



\vspace{-3pt}
Three-dimensional (3D) medical image segmentation under anisotropic voxel spacing (unequal spacing along the three axes, typically with substantially larger slice spacing in the z-direction than in the x and y directions) remains a core challenge in biomedical image analysis. Clinical MRI and CT scans frequently employ thick-slice acquisitions that create discontinuities along the z-axis. Standard 3D convolutional neural networks (CNNs) tend to overfit these sparsely sampled through-plane regions, while 2D CNNs, although robust in-plane, disregard volumetric context entirely. This imbalance between densely sampled intra-slice information and sparsely sampled inter-slice information has long constrained accurate volumetric delineation in clinical imaging workflows.

Recent hybrid designs have attempted to reconcile these regimes. Classical approaches such as 2.5D U-Net or nnU-Net \cite{isensee2021nnunet, isensee2024nnunetrevisited} mitigate anisotropy by empirically selecting 2D, 3D, or cascaded configurations per dataset. Yet these pipelines still rely on hand-crafted configuration rules or independent model ensembles. In contrast, MNet \cite{dong2022mnet} introduced a unified \textit{mesh} architecture that embeds 2D and 3D convolutions within each latent block, allowing the network to learn the optimal mixture of dimensional representations. Through its latent fusion of representation processes and multi-dimensional feature fusion, MNet balances inter- and intra-slice representations and demonstrated strong generalization across CT and MRI benchmarks including LiTS, KiTS, BraTS, and PROMISE \cite{lits,kits23,brats2020,promise12}.

Despite its conceptual elegance, reproducing MNet is non-trivial. The network’s combinatorial pathways and explicit, manually coded fusion operations (addition, subtraction, or concatenation) make it computationally heavy and potentially restrictive. Moreover, the original design lacks modern attention or state-space mechanisms that could enhance contextual reasoning along the z-direction. These limitations motivated our replication and extension effort to faithfully re-implement MNet within an nnU-Net-style framework and to explore whether learned fusion and lightweight long-range modelling can further improve anisotropy robustness. Our contributions are threefold:
\begin{enumerate}[topsep=0pt, itemsep=0.05pt, leftmargin=*]
    \item\textit{Reproduction:} We fully re-implemented MNet and validated its reported performance on the PROMISE and LiTS dataset, verifying the feasibility of its hybrid 2D/3D fusion strategy under realistic compute budgets.
    \item \textit{Fusion Gating:} We propose a dynamic gating mechanism that replaces hard-coded fusion choices with learned spatial or channel-wise gates, enabling the model to decide when and where to blend 2D and 3D features.
    \item \textit{VMamba Integration:} We augment MNet’s bottleneck stages with VMamba blocks, a state-space recurrent module that unfolds the feature map along the depth (z) axis, efficiently modelling long-range dependencies with O(D) complexity.
\end{enumerate}
We evaluate the reproduced and extended architectures on PROMISE and a controlled subset of the LiTS dataset, matched in size to ensure comparable statistical power. Experiments confirm the main claims of the original paper and show that our Fusion Gating and VMamba extensions yield consistent, modest gains in Dice similarity while reducing manual architectural decisions. Our findings reinforce MNet’s core premise, that adaptive 2D/3D fusion is key for anisotropic segmentation, and provide evidence that further automation through learned gating and state-space attention can improve both performance and reproducibility.

\section{Scope of Reproducibility}

The goal of this study is two fold: (1) verify the main empirical claims made by Dong et al.\cite{dong2022mnet} regarding the performance and robustness of MNet under anisotropic volumetric conditions. (2) Evaluate whether our proposed architectural extensions: Fusion Gating and VMamba integration, can further improve segmentation performance and anisotropy handling.


\subsection{Hypotheses.}
We define three testable hypotheses that guide our experimental design:
\begin{itemize}[topsep=0pt, itemsep=0.05pt, leftmargin=*]
    \item \textit{Hypothesis 1 (H1):} Our reproduced MNet implementation will match the performance trends reported in the original paper within an acceptable reproducibility margin ($\pm$2–3 Dice points).
    \item \textit{Hypothesis 2 (H2):} As inter-slice spacing increases from 1 to 4 mm, MNet and its extensions will exhibit measurable performance degradation, but the extended variants (Fusion Gating and VMamba) will show greater robustness (smaller Dice drop) compared to the reproduced MNet baseline.
    \item \textit{Hypothesis 3 (H3):} Introducing learned 2D–3D Fusion Gating and/or VMamba blocks will improve overall segmentation performance and training stability under optimal spacing conditions, relative to the reproduced MNet baseline.
\end{itemize}




\subsection{Replication Success Criteria.}
Reproduction is considered successful if our MNet implementation achieves Dice scores within approximately $\pm2-3$ percentage points of the values reported in the original paper, with consistent ranking among baselines. Extensions are considered successful if they consistently improve Dice scores and maintain stable training behaviour across both datasets without excessive computational overhead.


\section{Methodology}
\subsection{Dataset Description}
We evaluate all models on two publicly available benchmarks: PROMISE (MRI prostate segmentation) and a controlled subset of the LiTS dataset (CT liver and tumor segmentation). Both datasets were processed using the \textit{nnU-Net-v1} \cite{isensee2021nnunet} preprocessing pipeline to ensure consistency with the MNet authors released codebase, which internally relies on nnU-Net’s configuration logic and transformation suite.

\textbf{PROMISE12.} This dataset contains 50 training and 30 testing MRI volumes of the prostate with voxel spacings ranging approximately from 2.2 mm to 4.0 mm along the z-axis (median $\approx$ 3.6 mm). Each scan includes between 15 and 54 slices with in-plane dimensions up to 512×512 px. The task involves a single binary segmentation label (prostate). Preprocessing follows \textit{nnU-Net} defaults, including resampling to isotropic spacing where applicable, intensity normalization, cropping, and random augmentations (rotations, scalings, and elastic deformations) \cite{isensee2021nnunet}.

\textbf{LiTS Subset.} The full LiTS dataset comprises 131 CT scans from seven centers, with two labels (liver and liver tumor) and voxel spacings varying from 0.7 mm to 5.0 mm in the z-direction (median $\approx$ 1.0 mm). LiTS exhibits substantially finer native z-resolution than PROMISE. We sample 50 cases for training and 30 for validation to match PROMISE in size. Each scan typically contains 74–987 slices at 512×512 px resolution. Following MNet and nnU-Net preprocessing, each volume was resampled and intensity-normalized (HU clipping and z-score normalization), with identical augmentation policies applied. Two segmentation targets were retained: liver and tumor.


\textbf{Cross-Dataset Consistency.} For both datasets, we apply the same preprocessing, patch extraction, and five-fold cross-validation protocols to ensure differences arise from architecture rather than data handling. All augmentation, normalization, and sampling follow the nnU-Net framework, providing a consistent and reproducible setup for training and evaluation. To assess anisotropy robustness, we test three representative voxel spacings per dataset: PROMISE at 1 mm, 2.2 mm (optimal), and 4 mm; and LiTS at 1 mm (optimal), 2.0 mm, and 4 mm. These configurations mirror realistic acquisition variability and align with the original MNet study. Because the MNet repository relies on \textit{nnU-Net-v1}, we infer that the authors used five-fold cross-validation (80/20 split) rather than a fixed test set, which we replicate exactly.

\begin{figure}[t]
\centering
    \begin{subfigure}[b]{0.95\linewidth}
        \centering
        \includegraphics[width=0.85\linewidth, clip, trim=10 0 440 0]{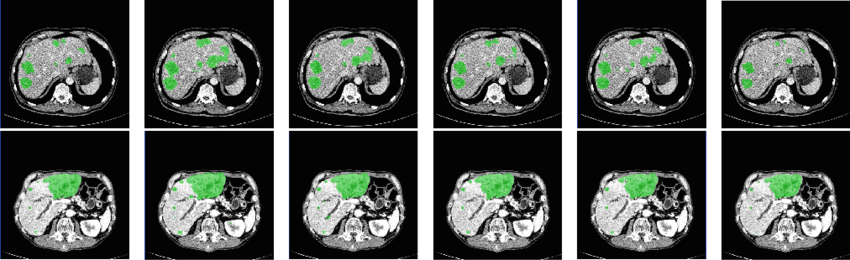}
        \caption{\label{fig:lits_ex} LiTS}
    \end{subfigure}
    \quad
    \begin{subfigure}[b]{0.95\linewidth}
        \centering
        \includegraphics[width=0.85\linewidth, clip, trim=50 200 650 50]{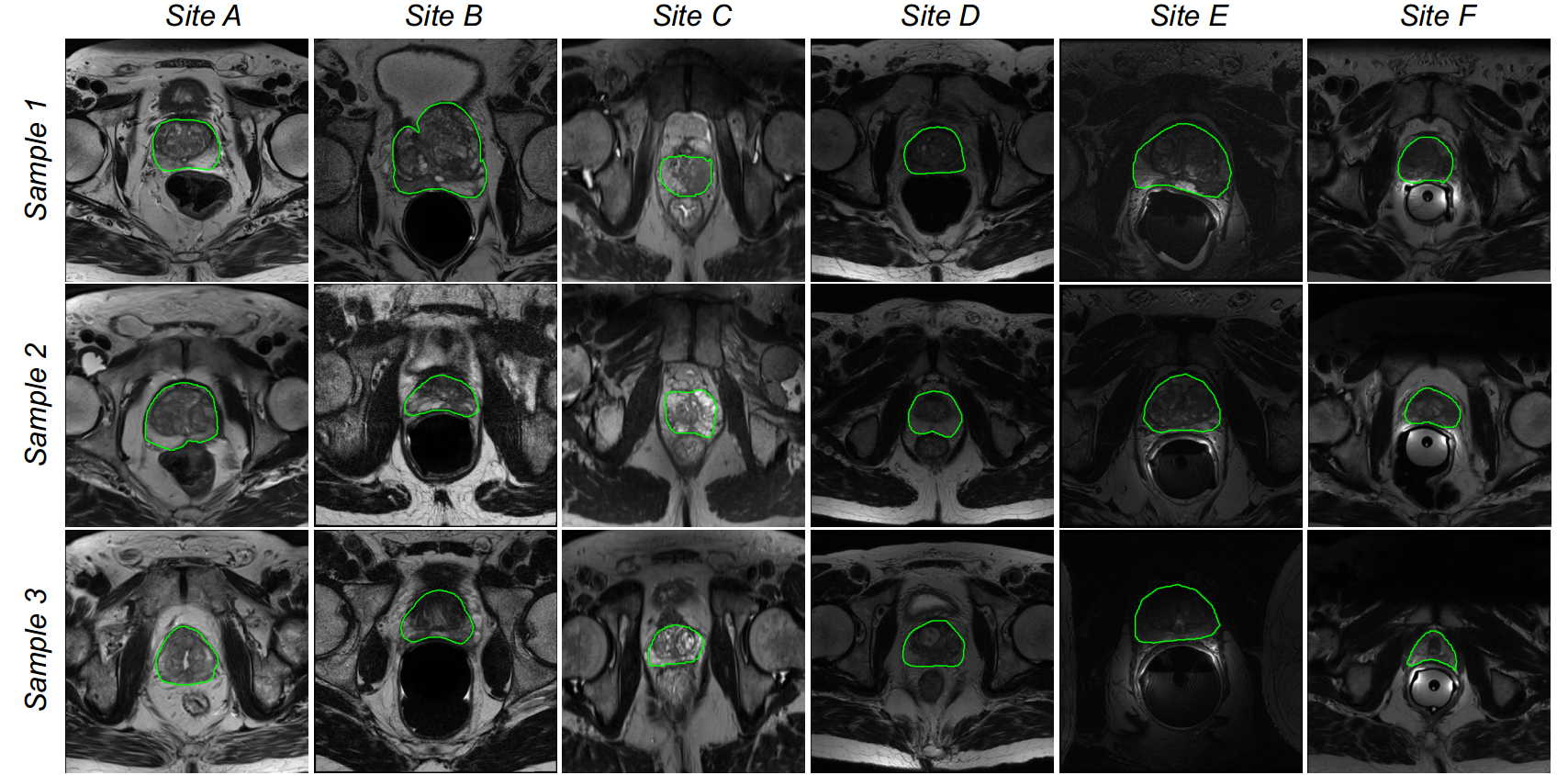}
        \caption{\label{fig:promise_ex} PROMISE}
    \end{subfigure}

    \caption{Representative examples illustrating anisotropy in PROMISE and LiTS. (a) LiTS CT (z-spacing-1.0 mm) with more uniform spacing. (b) PROMISE MRI (z-spacing-2.2 mm) showing fine in-plane but coarse inter-slice resolution \cite{lits,promise12}.}
    \label{fig:dataset_viz}
    \vspace{-15pt}
\end{figure}

\subsection{Model Description}
MNet \cite{dong2022mnet} is a hybrid 2D/3D convolutional network designed to address segmentation challenges in anisotropic medical images. Its central idea is to learn complementary representations from both slice-level (2D) and volumetric (3D) perspectives within a single framework, dynamically adjusting the contribution of each depending on voxel spacing and feature context.

\begin{figure}[t]
\centering
    \begin{subfigure}[b]{0.95\linewidth}
        \centering
        \includegraphics[width=0.7\linewidth]{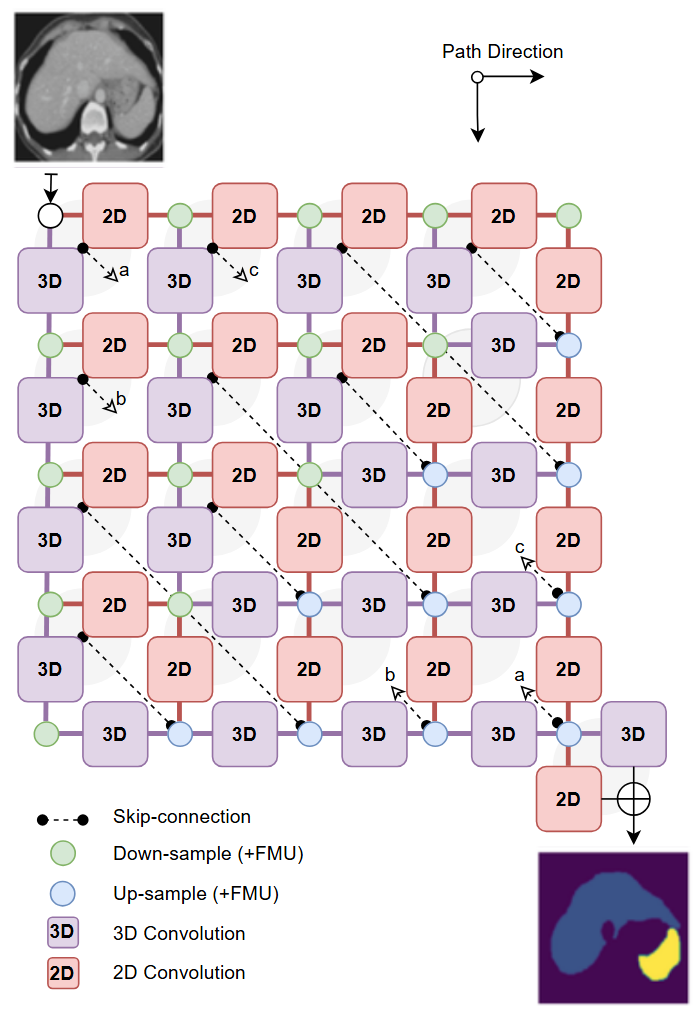}
        \caption{\label{fig:mnet_arch} MNet Architecture}
    \end{subfigure}
    \quad
    \begin{subfigure}[b]{0.95\linewidth}
        \centering
        \includegraphics[width=0.9\linewidth]{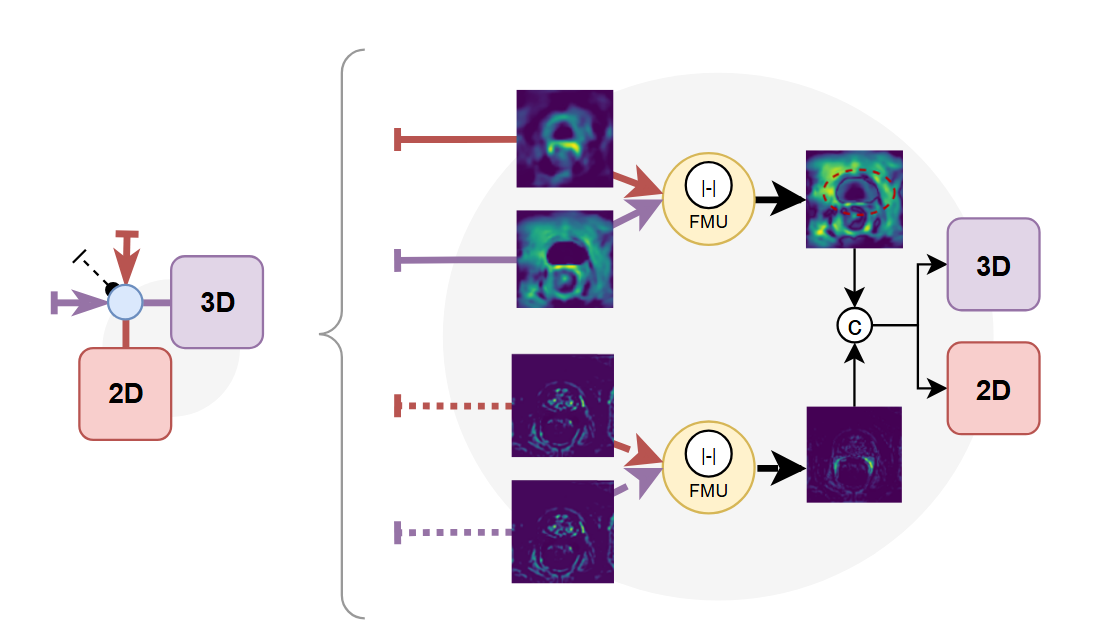}
        \caption{\label{fig:fmu_gate} Feature Merging Unit (FMU)}
    \end{subfigure}
    \caption{The architecture of our MNet. (a) The mesh structure makes the selections of representation processes unconstrained by embedding multi-dimensional convolutions deeply into latent basic modules. Supervision information is provided to six additional output branches to fully train shallow layers. (b) MNet latently fuses multi-dimensional and multi-level features inside basic modules, simultaneously taking the advantages of 2D and 3D representations, thus obtaining more accurate modelling for target regions \cite{dong2022mnet}.}
    \label{fig:schematic_mnet}
    \vspace{-10pt}
\end{figure}

\textbf{Architecture Overview.} MNet extends the U-Net family with parallel 2D and 3D encoder–decoder streams that operate on the same input volume. Each stream extracts spatial and contextual features at multiple scales. Their outputs are integrated through fusion gates, which combine features using addition, subtraction, or concatenation operations. This design allows MNet to interpolate smoothly between pure 2D, 2.5D, and full 3D behaviors, depending on the anisotropy of the input data. The network includes five encoder and five decoder stages, with symmetric skip connections between corresponding layers in both 2D and 3D paths to preserve fine details. Batch normalization and ReLU activations are applied after every convolutional block.

\textbf{Fusion Mechanism.} The fusion blocks are the defining component of MNet. Each block takes paired 2D and 3D feature maps and merges them through explicit arithmetic operations; elementwise addition, subtraction, or concatenation, followed by a convolutional refinement. This \textit{manual} gating determines how spatial and volumetric information interact at each resolution level. While this offers interpretability and stability, it also introduces rigidity, motivating our later Fusion Gating extension that enables learnable weighting instead of fixed fusion rules.

\textbf{Implementation Details.}
We reproduced the architecture using PyTorch, following the configuration and hyperparameters described in the official MNet GitHub implementation \cite{dong2022mnet}. The model is trained with the Dice loss, optimized via Adam with an initial learning rate of 1e–4 and a cosine annealing schedule. Mixed precision training is enabled to reduce GPU memory consumption. Parameter count for the baseline model is approximately 8.77 M, consistent with the original paper. All experiments use the nnU-Net v1 infrastructure for data loading, preprocessing, and augmentation, ensuring procedural alignment with the original setup.

\textbf{Extensions} While MNet's fixed fusion rules work well on average, they cannot adapt to local variations in image quality or anisotropy severity. Similarly, standard 3D convolutions lack the global receptive field needed for coherent depth modelling in thick-slice scans. To address these limitations, we propose two targeted extensions: Fusion Gating and VMamba Blocks. It should be noted that all extensions were applied only to our re-implementation and not to the original MNet codebase. Therefore, if improvements are observed relative to our version of MNet, we can reasonably expect similar improvements in the original model, assuming the reproduction results are comparable.

\subsection{Fusion Gating Extension}
The original MNet’s fusion blocks (Feature Merging Units, FMUs) combine 2D and 3D feature maps using one of three fixed operations: elementwise sum, subtraction, or concatenation. While effective, these static rules assume a uniform mixing ratio across all spatial locations and channels. Consequently, they cannot adaptively emphasize the more reliable feature stream in different anatomical regions or suppress modality-specific noise. In addition, concatenation increases channel dimensionality and computational load downstream. To address these limitations, we introduce \textit{Fusion Gating}, a learned, data-driven mechanism that determines how much information to take from each feature stream (2D or 3D) per channel or per voxel. This allows the model to adaptively modulate between slice-level and volumetric cues, improving robustness to variable anisotropy.

\begin{figure}[t]
    \centering
    \captionsetup{width=0.95\linewidth}
    \includegraphics[width=0.95\linewidth]{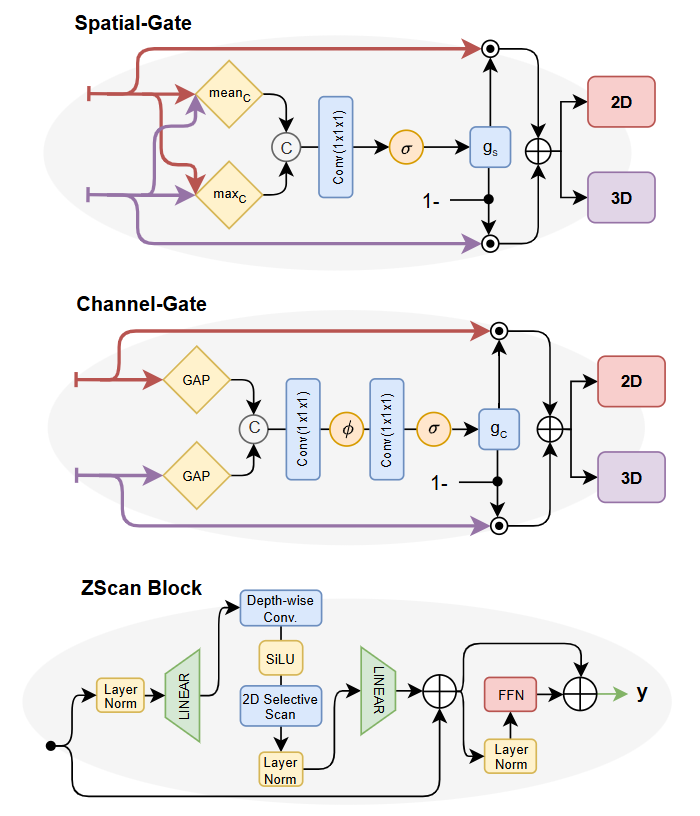}
    \caption{Overview of Fusion Gating variants and VMamba ZScan schematic. Given 2D and 3D feature maps at a matching scale, the gating module predicts either channel-wise or voxel-wise trust maps to interpolate between the two representations. The learned gates replace static fusion rules, allowing data-adaptive feature mixing.}
    \label{fig:mnet_improvements}
    \vspace{-12pt}
\end{figure}

\textbf{Mechanism.} 
Let $x^{2D},x^{3D}\in\mathbb{R}^{N\times C\times D\times H\times W}$ be the 2D and 3D feature tensors at a corresponding resolution, where $N$ is the batch size, $C$ the number of channels, and $D\times H\times W$ the spatial dimensions. Fusion Gating learns a soft gate $g \in [0,1]^{N\times C\times D\times H\times W}$ that interpolates between the two streams:
\begin{equation*}
y = g \odot x^{2D} + (1 - g) \odot x^{3D}
\end{equation*}
Where $\odot$ denotes element-wise multiplication. The gate $g$ may be computed in one of two modes:
\begin{enumerate}[topsep=0pt, itemsep=0.10pt, leftmargin=*]
    \item \textit{Channel gate:} A global confidence score is estimated for each channel, independent of spatial location.
 \item \textit{Spatial gate:} Here the gate is computed for each spatial location, shared across channels.
 \end{enumerate}
The a schematic of the fusion gating variants can be seen Figure~\ref{fig:mnet_improvements}, whereas a detailed overview of the explicit mechanism break downs for fusion gating can be found in Appendix \ref{app:B}.

\textbf{Integration within MNet.} Fusion Gating replaces the original FMUs at points where both 2D and 3D feature streams are available. In the encoder, gating occurs after pooling; in the decoder, it operates on both the skip connections and upsampled features. Purely 2D or 3D pathways bypass gating to preserve computational efficiency.

\textbf{Computation and Stability.} Channel gating adds two $1\times 1\times 1$ convolutional layers ($\approx140$k parameters in total), while spatial gating adds a single $1\times 1\times 1$ layer, resulting in negligible parameter overhead. Both variants remain stable during training due to bounded sigmoid activations and neutral initialization.

\subsection{VMamba Block Extension} 
Anisotropic medical scans often exhibit large disparities between in-plane and through-plane resolution, limiting the effectiveness of conventional 3D convolutions in modelling long-range context along the depth (z) axis. While attention-based methods could provide global dependencies, they are computationally expensive for volumetric data. To address this, we developed a VMamba-based approach where a \textit{z-axis selective scan} is implemented that efficiently captures inter-slice dependencies with O(D) complexity per spatial position \cite{liu2024vmamba_arxiv}. 

\textbf{Mechanism:} The approach inserts z-axis VMamba blocks (\textit{CBzMamba}) into bottleneck stages 3, 4, and 5 of the MNet encoder. Each spatial position $(H,W)$ is processed independently as a sequence along depth D. The input volume is reshaped from ($N,C,D,H,W$) to ($N \cdot H \cdot W, D, C$), treating each of the $N \cdot H \cdot W$ spatial positions as an independent sequence of length D. The selective state-space model then processes these sequences in parallel:
\begin{equation*}
    \mbox{\footnotesize $\displaystyle
    \text{CBzM}(x) = \text{Conv}_{1\times1\times1}^{\text{reduce}}(x)
    \rightarrow \text{ZScan}(x) \rightarrow \text{Conv}_{1\times1\times1}^{\text{expand}}(x)
    $}
\end{equation*}
where ZScan performs the z-axis selective state-space scan:
\begin{equation*}
\mbox{\footnotesize $\displaystyle
x' = \text{ZScan}(x) = f_{\text{SSM}}(\text{reshape}(x; (N \cdot H \cdot W), D, C)) 
$}
\end{equation*}
This formulation processes each ($H,W$) position's depth sequence independently, achieving O(D) complexity per position with total cost O($D \cdot H \cdot W$). The approach remains highly parallelizable across spatial positions, making it memory-efficient for anisotropic volumes where D is typically small (16–32 slices at bottleneck stages).

\textbf{Parameterization.} Each CBzMamba block first reduces channels by a factor $r=0.5$, applies the state-space scan, and restores dimensionality. The combined VMamba architecture totals 7.42M parameters, a 15\% reduction from the baseline 8.77M, as efficient state-space operations replace heavier 3D convolutions in the bottleneck stages.

\textbf{Benefits for Anisotropy.} The z-axis scanning provides effective global receptive fields across slices, improving feature coherence and boundary continuity in anisotropic volumes, especially when slice spacing is coarse (2–4mm) compared to in-plane resolution (0.5–1mm). By explicitly modeling depth dependencies, VMamba helps the decoder produce globally consistent predictions without losing in-plane precision.

\textbf{Computation and Stability.} Z-axis VMamba operates with O(D) complexity per spatial position and scales efficiently due to its insertion at low-resolution bottleneck stages. GPU memory usage is comparable to (or lower than) stacked 3D convolutions. Training stability is strong when applied only at deep semantic layers; early-stage application may slow convergence due to high spatial dimensions.

\textbf{Complementarity with Fusion Gating.} Fusion Gating and VMamba address distinct architectural dimensions: the former governs inter-stream fusion between 2D and 3D pathways, while VMamba enhances intra-stream coherence along z within the 3D path. In combination, they provide additive gains by refining both the fusion process and depth-level feature semantics.

\section{Training and Evaluation}

Three primary experimental groups were conducted for each dataset:
\begin{enumerate}[topsep=0pt, itemsep=0.25pt, leftmargin=*]
    \item \textit{Baseline comparison:} Replication of the original MNet (150 epochs) and re-implementation with proposed extensions (Fusion Gating, VMamba; each 150 epochs) at optimal spacings (PROMISE = 2.2 mm, LiTS = 1.0 mm). 
    \item \textit{Anisotropy study:} Controlled z-spacing experiments (1 mm, 2.2/2.0 mm, 4 mm) over 50 epochs using a single 80/20 split.
    \item \textit{Ablation study:} Short 50-epoch runs at optimal spacing, evaluating each innovation individually (Channel Gate, Spatial Gate, VMamba) before selecting the best variants for full-scale comparison.
\end{enumerate}

All experiments were executed on \textit{Lightning.ai} cloud infrastructure using NVIDIA L4 GPUs (16 GB VRAM) and a 4-core CPU (64 GB RAM). Due to limited compute compared to the MNet authors, experiment counts and epochs were pragmatically constrained to 50/150 epochs, while ensuring sufficient runs for reproducibility. This decision is justified by: (1) convergence analysis showing Dice plateaus by epoch 100-120 (Appendix~\ref{app:A}) for the PROMISE dataset, (2) our 150-epoch baseline achieving 89.1 \%, only 0.7 \% below the 500-epoch result, and (3) standard reproducibility criteria accepting ±2-3 \% tolerance. The minimal performance gap confirms that extended training beyond convergence does not materially affect our conclusions on the PROMISE dataset. While it was hoped that the LiTS dataset would exhibit the same trends, this was not as evident. The loss curves showed increasingly erratic behaviour, indicating that additional data or longer training times would likely be the preferred approach. However, this was not feasible within the scope of the current work. A summary of the computational details are shown in Table~\ref{tab:compute_resources}.  

\renewcommand{\arraystretch}{1.1}
\begin{table}[t]
\centering
\captionsetup{width=0.95\linewidth}
\footnotesize
\resizebox{\columnwidth}{!}{%
\begin{tabular}{p{3cm}p{3.5cm}p{3cm}}
\toprule
\textbf{Resource} & \textbf{Configuration} & \textbf{Notes} \\ 
\midrule
GPU & NVIDIA L4 (16\,GB VRAM) & Lightning.ai cloud \\
CPU / RAM & 4 cores / 64\,GB & Shared host \\
Epoch runtime & 225--320\,s & Dataset/architecture dependent \\
Epochs & 50 (short) / 150 (full) & Fixed per design \\
AMP / Checkpointing & Enabled & via \textit{nnU-Net-v1} \\
Total GPU hours & $\approx$ 150--175 & Single dataset \\
\bottomrule
\end{tabular}
}
\vspace{5pt}
\caption{Computational resource summary.}
\label{tab:compute_resources}
\vspace{-15pt}
\end{table}

\renewcommand{\arraystretch}{1.10}
\begin{table}[t]
\centering
\captionsetup{width=0.95\linewidth}
\footnotesize
\resizebox{\columnwidth}{!}{%
\begin{tabular}{p{2.5cm}p{3.5cm}p{3.5cm}}
\toprule
\textbf{Aspect} & \textbf{Setting / Value} & \textbf{Notes} \\ 
\midrule
Validation strategy & 5-fold CV (3 folds reported) & Consistent random seeds \\
Split ratio & 80/20 & Single split for ablation/anisotropy \\
Primary metric & Dice Similarity Coefficient (DSC) & Volumetric overlap \\
Averaging & Mean $\pm$ std across folds & Per dataset \\
\bottomrule
\end{tabular}
}
\vspace{5pt}
\caption{Evaluation configuration and metrics.}
\label{tab:evaluation}
\vspace{-20pt}
\end{table}

\subsection{Loss Description}
Training follows the original MNet hybrid objective combining Dice and Cross-Entropy under deep supervision. Six auxiliary output branches are attached via $1\times1\times1$ convolutions at multiple decoder depths, each producing a resampled prediction. The final objective is a weighted sum:
\begin{equation*}
    \mbox{\footnotesize $\displaystyle
    \begin{aligned}
        \mathcal{L} &= \mathcal{L}(X_{55},Y_{55}) + \sum_{i=2}^{4}\lambda_i[\mathcal{L}(X_{5i},Y_{5i})+\mathcal{L}(X_{i5},Y_{i5})] \\
        \lambda_i &= (\sfrac{1}{2})^{5-i}
    \end{aligned}
    $}
\end{equation*}
where $X_{ij}$ denotes the output from encoder stage $i$ and decoder stage $j$, and $Y_{ij}$ is the corresponding downsampled ground truth.

Optimization used Stochastic Gradient Descent (SGD) ($lr = 1 \times 10^{-2}$ and $momentum=0.99$) with polynomial decay scheduling \cite{chen2018deeplab}. Batch and patch sizes followed nnU-Net’s automatic configuration to ensure consistent GPU utilization across 2D and 3D branches.

\subsection{Evaluation Protocol}
Evaluation used identical preprocessing, augmentation, and normalization as training to isolate architectural differences. Data augmentation includes rotation, scaling, elastic deformation, and oblique-plane transformations., which is automatically handled by the nnU-Net framework \cite{isensee2021nnunet}. Cross-validation (5-fold; mean of 3 folds reported) was used for full experiments, while single 80/20 splits were applied in ablation and anisotropy tests. The primary performance metric is the \textit{Dice Similarity Coefficient (DSC)}:
\begin{equation*}
    \mbox{\footnotesize $\displaystyle
        \mathrm{DSC} = \frac{2|X \cap Y|}{|X| + |Y|}
    $}
\end{equation*}
where $X$ and $Y$ are predicted and reference voxel sets. Mean $\pm$ standard deviation values are reported across folds; evaluation settings are summarized in Table~\ref{tab:evaluation}.



\section{Results}
\subsection{Main Reproduction}

Table~\ref{tab:baseline_comparison} summarizes the primary reproduction results. On PROMISE12, our re-implementation achieves 89.0 $\pm$ 0.9\% Dice, which is within 0.8 points of the originally reported 89.8\% and within the accepted reproducibility margin of approximately 2 to 3 percent. For LiTS, the reproduced model reaches 94.3 $\pm$ 1.9\% for liver and 54.6 $\pm$ 3.1\% for tumor segmentation. This is consistent with the original liver result of 94.3 percent, while the lower tumor Dice is expected given our smaller training subset of 50 cases compared to the full 131 cases and a shorter training schedule of 150 epochs instead of 500. PROMISE12 training curves show smooth and stable convergence. LiTS curves exhibit higher oscillation due to the combination of class imbalance, image heterogeneity, and reduced sample count. See Appendix \ref{app:A}, for all corresponding training loss curves for each experiment and cross-validation fold.

\begin{figure}[t]
\centering
\captionsetup{width=0.95\linewidth}
    \begin{subfigure}[b]{\linewidth}
        \centering
        \includegraphics[width=\linewidth]{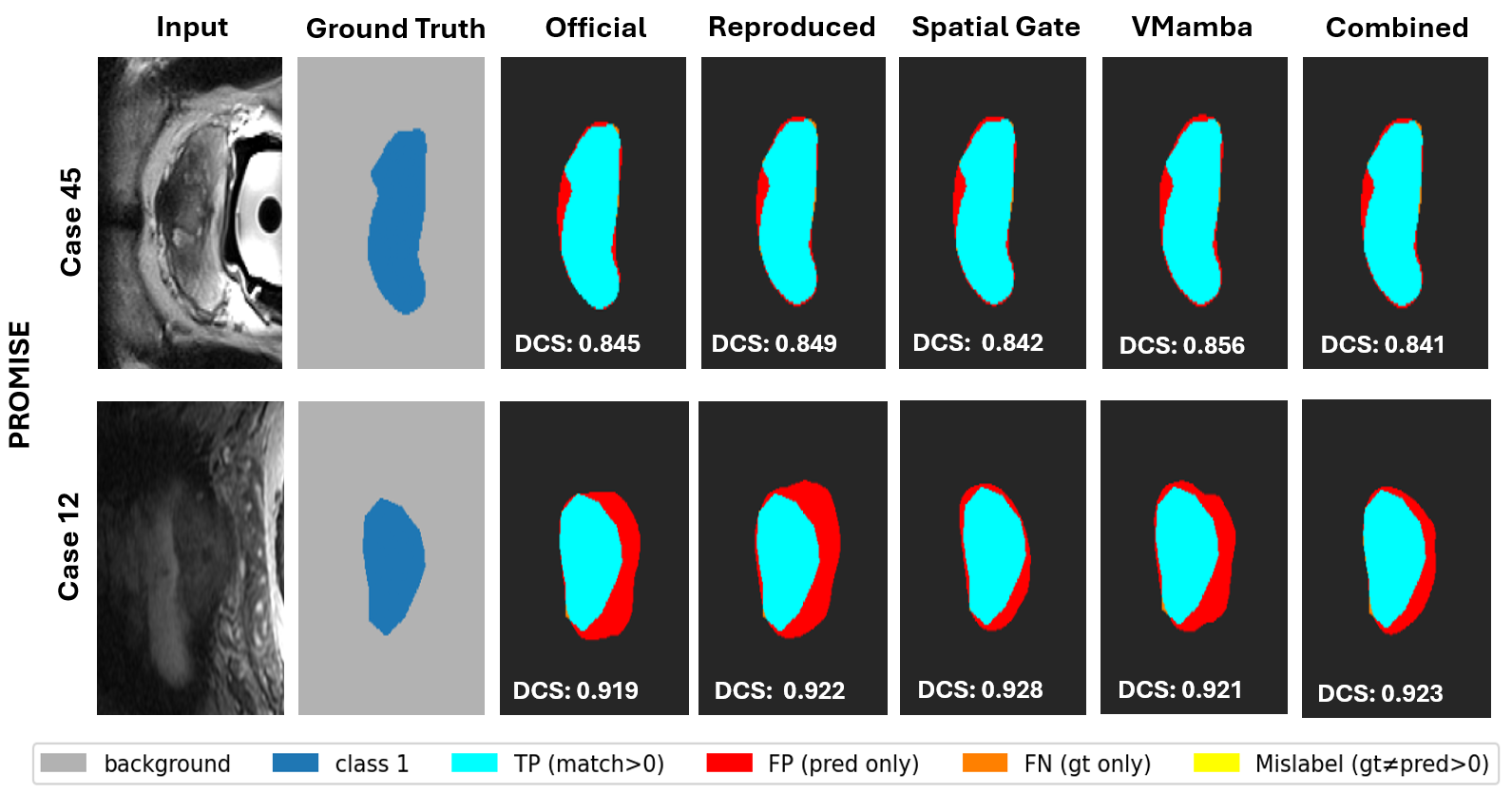}
        \caption{\label{fig:promis_seg} PROMISE}
    \end{subfigure}
    \quad
    \begin{subfigure}[b]{\linewidth}
        \centering
        \includegraphics[width=\linewidth]{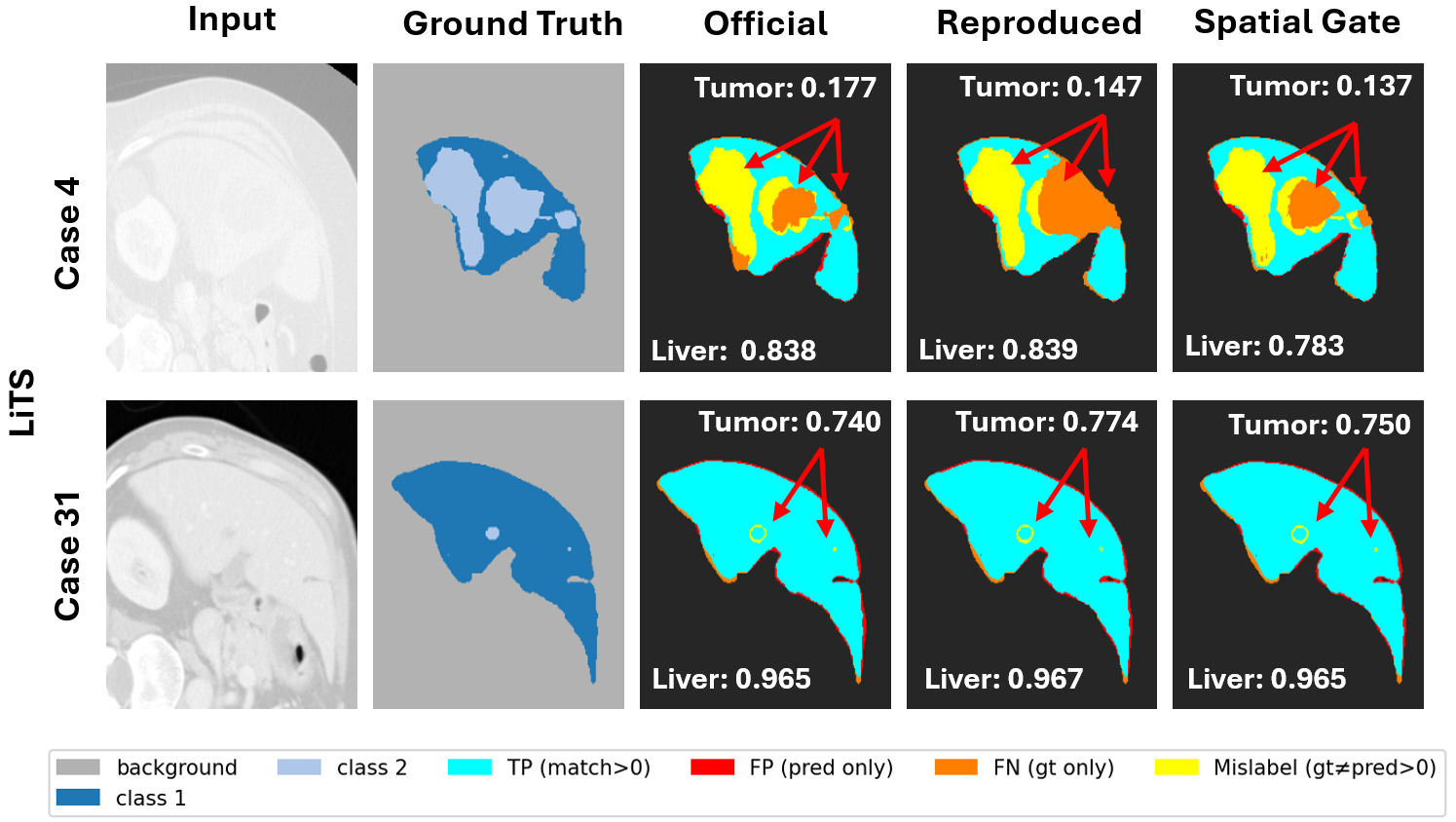}
        \caption{\label{fig:lits_seg} LiTS}
    \end{subfigure}

    \caption{Qualitative comparison of segmentation performance across datasets. The figure presents visual results on two benchmark datasets: PROMISE (prostate MRI; top) and LiTS (liver and tumor CT; bottom). For each dataset, two randomly selected unseen test cases are shown. Each case includes the input image, the ground-truth segmentation mask, and the corresponding predicted masks from different experimental configurations: Official, Reproduced, Spatial Gate, VMamba, and Combined (SG + VM) for 150 epochs. The overlay maps display correctly segmented regions: True Positives (cyan), False Positives (red), False Negatives (orange), and mislabeled regions (yellow). The Dice Similarity Coefficient (DCS) for each prediction is provided above each example for quantitative reference. All visualizations are cropped and zoomed during preprocessing for clarity of the target structures.}
    \label{fig:segmentation_image}
    \vspace{-15pt}
\end{figure}

\subsection{PROMISE Qualitative Analysis}

Figure~\ref{fig:segmentation_image} (top) shows representative PROMISE examples. In Case 45, all variants including the official model, our reproduction, Spatial Gate, VMamba, and the combined model produce visually consistent and accurate prostate masks. This matches the very small performance differences reported in Table~\ref{tab:baseline_comparison}, where all variants fall within approximately 0.5 to 1.0 percent of each other. In Case 12, the Spatial Gate and combined variants better follow the prostate boundary and reduce peripheral false negatives, consistent with their modest improvement over the baseline model. Overall, all methods demonstrate stable performance on PROMISE and maintain reliable gland localization even under mild anisotropy at 2.2mm.

\subsection{LiTS Qualitative Analysis}

Figure~\ref{fig:segmentation_image} (bottom) presents two LiTS cases that illustrate typical and difficult scenarios. Case 31 contains well defined tumors with good contrast. All variants accurately segment the liver and identify the tumor regions, which agrees with the ablation results in Table~\ref{tab:ablation} where tumor Dice scores reach as high as 63.8 percent under favourable conditions. Case 4 highlights the primary challenge of LiTS. This volume contains diffuse, low contrast, and partially occluded tumor regions. All models struggle to recover these regions and predictions become fragmented or incomplete. This case dependent difficulty explains the higher variance reported in Table~\ref{tab:baseline_comparison}, particularly for VMamba variants that show both the highest mean tumor Dice and the highest standard deviation. Liver segmentation remains consistently strong across all models, whereas tumor performance strongly depends on lesion morphology, dataset size, and visibility.

\subsection{Observed Challenges and Interesting Outcomes}

Two consistent observations emerge from the results.
\begin{enumerate}[topsep=0pt, itemsep=0.25pt, leftmargin=*]
    \item \textit{LiTS Data Limitations:} Tumor segmentation is primarily limited by data complexity rather than architectural choices. All models achieve liver Dice above 94 percent, but tumor Dice remains between 52 and 55 percent across folds. VMamba slightly improves average tumor Dice but also shows increased variance. This behaviour reflects tumor sparsity (class imbalance), irregular shapes, and limited training volume rather than instability within the architecture.
    \item \textit{Extensions are complementary, not disruptive:} Both Spatial Gate and VMamba improve stability without disrupting the original MNet behaviour. Spatial Gating improves boundary precision and produces small but consistent gains on PROMISE. VMamba improves depth continuity and yields the most stable liver performance with standard deviations between 0.3 and 0.7 percent. Together, these results confirm that improved 2D to 3D fusion and enhanced z axis context modelling complement the original design.
\end{enumerate}

\renewcommand{\arraystretch}{1.05}
\begin{table*}[t]
\centering
\captionsetup{width=0.95\linewidth}
\scriptsize
\begin{subtable}[t]{\textwidth}
\centering
\setlength{\tabcolsep}{4.5pt}
\resizebox{\textwidth}{!}{%
\begin{tabular}{p{2.5cm}p{1.5cm}p{1.5cm}p{2cm}p{2.5cm}p{2cm}p{2cm}}
\toprule
\textbf{Model} & \textbf{\#Params (M)} & \textbf{Epochs} & \textbf{CV Folds} & \textbf{PROMISE (Prostate)} & \textbf{LiTS (Liver)} & \textbf{LiTS (Tumor)} \\
\midrule
3D U-Net (Original \cite{dong2022mnet}) & --- & --- & --- & 85.6 & 90.1 & 51.0 \\
nnU-Net (Original \cite{dong2022mnet}) & --- & --- & --- & 89.5 & 94.1 & 62.0 \\
MNet (Original \cite{dong2022mnet}) & 8.77 & 500 & --- & 89.8 & 94.3 & 66.3 \\ \hline
MNet (Original) & 8.77 & 150 & 3 (80/20) & 89.1 $\pm$ 0.7 & 94.4 $\pm$ 2.0 & 52.1$\pm$3.9 \\
MNet (Ours) & 8.77 & 150 & 3 (80/20) & 89.0 $\pm$ 0.9 & 94.3$\pm$1.9 & 54.6 $\pm$ 3.1 \\
MNet + Spatial Gate & 8.77 & 150 & 3 (80/20) & \textbf{89.2} $\pm$ \textbf{1.0} & 94.4$ \pm $1.7 & 52.7$\pm$2.8 \\
MNet + VMamba & 7.42 & 150 & 3 (80/20) & 88.9 $\pm$ 1.2  & 95.8 $\pm$ 0.7 & \textbf{55.0} $\pm$ \textbf{10.8} \\
MNet + (SG + VM) & 7.42 & 150 & 3 (80/20) & \text{89.0} $\pm$ \text{1.5} & \textbf{95.9} $\pm$ \textbf{0.3} & 53.7 $\pm$ 13.6 \\
\bottomrule
\end{tabular}

}
\caption{Baseline comparison ($\text{Dice \% mean}\,\pm \,\text{std})$ over evaluated folds.}
\label{tab:baseline_comparison}
\end{subtable}
\vspace{6pt} 

\begin{subtable}[t]{\textwidth}
\centering
\setlength{\tabcolsep}{4.5pt}
\resizebox{\textwidth}{!}{%
\begin{tabular}{p{2.75cm}p{1.75cm}p{2.25cm}p{2.75cm}p{2.25cm}p{2.25cm}}
\toprule
\textbf{Model} & \textbf{\#Params (M)} & \textbf{CV / Epochs} & \textbf{PROMISE (Prostate)} & \textbf{LiTS (Liver)} & \textbf{LiTS (Tumor)} \\
\midrule
MNet (Ours) & 8.77 & 1 (80/20) / 50 & 88.6 & 94.30 & 60.09 \\
MNet + Channel Gate & 8.91  & 1 (80/20) / 50 & 89.8 & 92.72 & 63.26 \\
MNet + Spatial Gate & 8.77  & 1 (80/20) / 50 & \textbf{90.0} & 92.53 & 54.86 \\
MNet + VMamba & 7.42  & 1 (80/20) / 50 & 88.8 & \textbf{94.80} & 55.89 \\
MNet + (CG + VM) & 7.42  & 1 (80/20) / 50  & 89.1 & 94.52 & \textbf{63.80} \\
MNet + (SG + VM) & 7.42  & 1 (80/20) / 50  & 89.7 & 94.07 & 56.78 \\
\bottomrule
\end{tabular}
}
\caption{Ablation study at optimal spacing (PROMISE = 2.2 mm, LiTS = 1.0 mm).}
\label{tab:ablation}
\end{subtable}
\vspace{6pt} 

\begin{subtable}[t]{\textwidth}
\centering
\setlength{\tabcolsep}{4.5pt}
\resizebox{\textwidth}{!}{%
\begin{tabular}{p{2cm}p{1.5cm}p{0.90cm}p{1.75cm}p{2.7cm}p{2.65cm}p{2.65cm}}
\toprule
\textbf{Model} & \textbf{\#Params (M)} & \textbf{CV / Ep.} & \textbf{Z-space (mm)} & \textbf{PROMISE (Prostate)} & \textbf{LiTS (Liver)} & \textbf{LiTS (Tumor)} \\
\midrule
MNet (Original) & 8.77 & 1 / 50 & 1.0, 2.2, 4.0 & 89.6, 89.3, \textbf{88.5} (89.1) & \multicolumn{2}{c}{80.30, 82.82, 79.13 \cite{dong2022mnet}} \\
MNet (Ours) & 8.77 & 1 / 50 & 1.0, 2.2, 4.0 & 89.5, 88.6, 87.5 (88.5) & 91.4, 95.0, 93.8 (93.4) & 61.3, 52.9, 50.7 (55.0) \\
MNet + Spatial Gate & 8.77 & 1 / 50 & 1.0, 2.2, 4.0 & 89.6, \textbf{90.0}, 88.1 (\textbf{89.2}) & 93.7, 94.8, 92.6 (93.7) & 65.0, 61.8, 50.7 (59.2) \\
MNet + VMamba & 7.42 & 1 / 50 & 1.0, 2.2, 4.0 & \textbf{89.9}, 88.8, 88.2 (89.0) & \textbf{94.7}, \textbf{95.1}, 92.5 (\textbf{94.1}) & \textbf{65.2}, 61.1, 52.1 (\textbf{59.5}) \\
MNet + (SG + VM) & 7.42 & 1 / 50 & 1.0, 2.2, 4.0 & 89.4, 89.7, 87.6 (88.9) & 94.1, \textbf{95.1}, 91.1 (93.4) & 62.9, \textbf{62.2}, 52.4 (59.2) \\
\bottomrule
\end{tabular}
}
\caption{Anisotropy sensitivity (Dice \%) across z-spacings (1 mm, 2.2/2.0 mm, 4 mm). ($\cdot$) indicates the mean across the Z-space.}
\label{tab:anisotropy}
\end{subtable}
\caption{Results across PROMISE and LiTS. Each sub-table reports Dice (\%) for a specific analysis.}
\label{tab:all_results}
\vspace{-10pt}
\end{table*}

\subsection{Extensions and Ablations}

\textbf{Fusion Gating.}  
Table~\ref{tab:ablation} shows that Spatial Gating provides the strongest improvement on PROMISE, achieving 90.0 percent Dice with minimal parameter overhead. On LiTS, Spatial Gating maintains competitive liver performance in the range of 92.5 to 94.8 percent and produces stable tumor Dice between 54.8 and 61.8 percent. Channel Gating was only partially explored due to time and computational limits.

\textbf{VMamba.}  
Replacing the bottleneck convolutions with VMamba reduces parameters from 8.77 million to 7.42 million and produces some of the best liver results in both baseline and ablation settings. VMamba reaches 95.8 $\pm$ 0.7 percent liver Dice in the 150 epoch setting and 94.8 percent in the shorter ablation run. Tumor Dice shows a small improvement on average but higher fold to fold variance, reflecting sensitivity to lesion heterogeneity.

\textbf{Combined VMamba and Spatial Gate.}  
The combined variant achieves strong liver results, reaching 95.9 $\pm$ 0.3 percent, and performs comparably to or better than the individual components. Tumor Dice remains limited by dataset complexity rather than fusion or state space modeling. Additional tuning or extended training schedules may be required to leverage the complementarity of the two modules.

\subsection{Anisotropy Sensitivity}

Table~\ref{tab:anisotropy} evaluates model robustness to changes in inter slice spacing. On PROMISE, the Spatial Gate model is the most stable, decreasing only from 89.6 percent at 1 millimeter to 88.1 percent at 4 millimeters, a drop of approximately 1.5 points. This is smaller than the baseline model, which drops from 89.5 to 87.5 percent. VMamba also maintains stable performance with Dice decreasing from 89.9 percent to 88.2 percent across the same range. On LiTS, all variants follow a similar trend. Liver Dice remains high across all spacings while tumor Dice decreases more sharply with increasing spacing. The Spatial Gate, VMamba, and combined variants produce similar robustness profiles, confirming that both learned fusion and depth aware modeling help preserve MNet performance under increasing anisotropy. In the case of the LiTS dataset, it should be noted that due to compute limitations, a full original evaluation could not be completed. Instead, the original results reported in the MNet paper \cite{dong2022mnet} were used as a reference. Unfortunately, this benchmark uses 500 epochs and only  provides the combined mean scores, but it still serves as a valid basis for visual comparison.

\begin{figure}[t]
\centering
\captionsetup{width=0.95\linewidth}
    \begin{subfigure}[b]{\linewidth}
        \centering
        \includegraphics[width=0.85\linewidth]{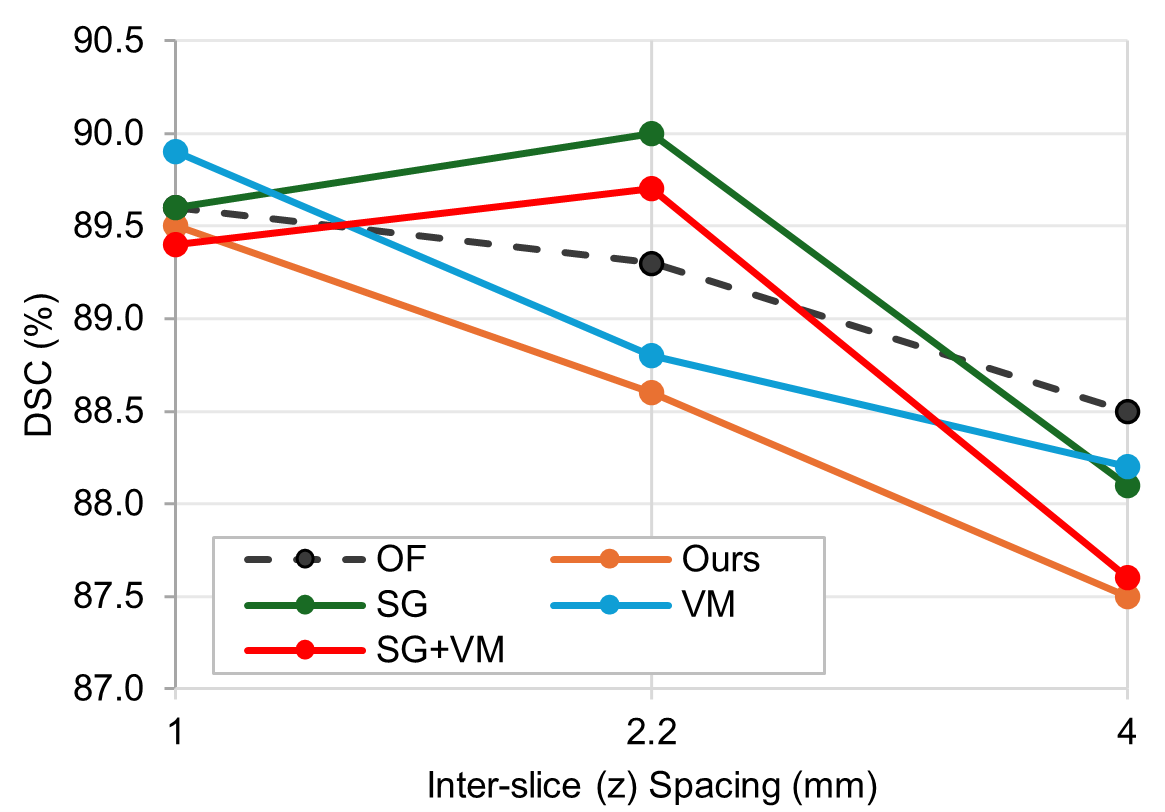}
        \caption{PROMISE}
        \label{fig:promise_anisotropy}
    \end{subfigure}
    \quad
    \begin{subfigure}[b]{\linewidth}
        \centering
        \includegraphics[width=0.85\linewidth]{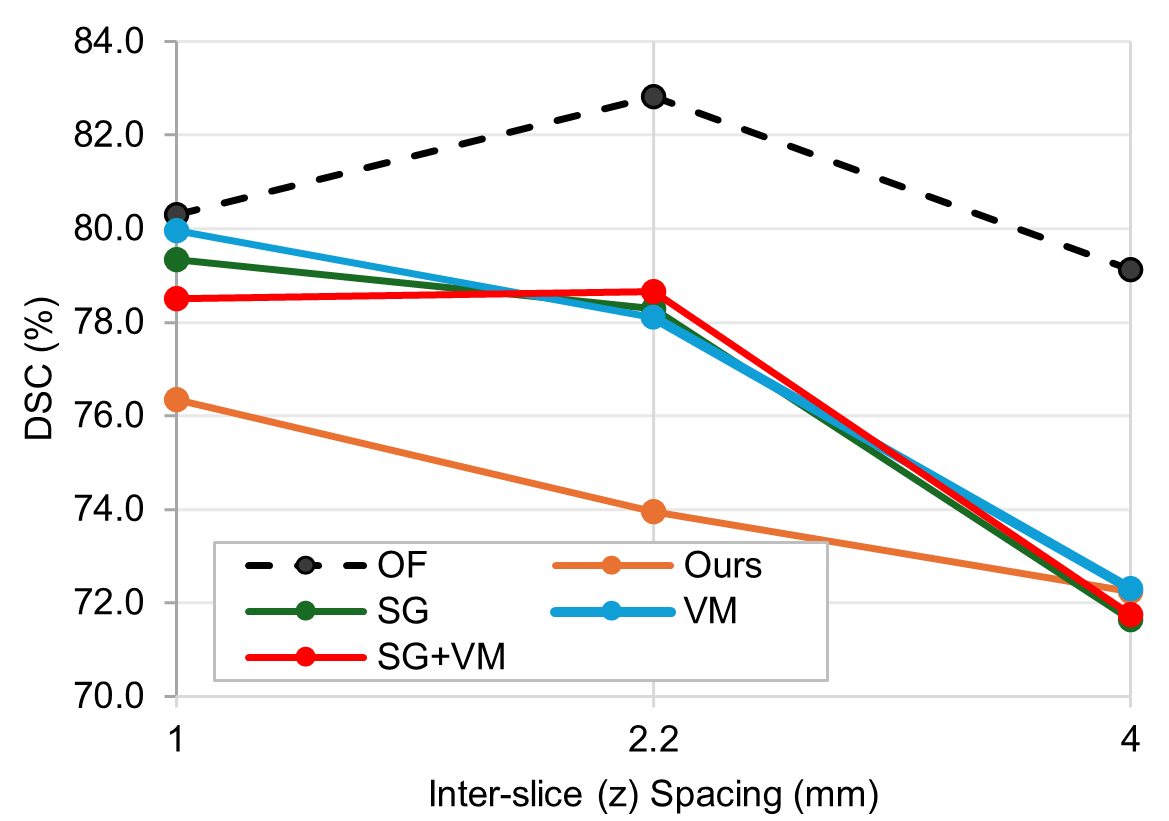}
        \caption{LiTS}
        \label{fig:lits_anisotropy}
    \end{subfigure}
    \caption{Anisotropy comparison investigation where mean Dice is used as the evaluation metric across classes.}
\label{fig:anisotropy_image}
\vspace{-15pt}
\end{figure}




\section{Conclusion}
Our study successfully reproduced MNet’s core performance and demonstrated that simple, lightweight extensions can further improve anisotropic segmentation stability. The Spatial Fusion Gate yielded consistent gains with negligible overhead, while VMamba enhanced depth continuity and reduced parameters, albeit requiring longer training to converge fully. Table~\ref{tab:hypotheses} summarizes how our results relate to the three predefined hypotheses.

\renewcommand{\arraystretch}{1.4}
\begin{table}[t]
\centering
\scriptsize
\captionsetup{width=0.95\linewidth}
\resizebox{\columnwidth}{!}{%
\begin{tabular}{m{0.4cm} m{6.6cm}}
\hline
\textbf{ID} & \makecell[c]{\textbf{Hypothesis and Outcome Summary}} \\
\hline

\textbf{H1} & 
\begin{minipage}[t]{\linewidth}
\begin{itemize}[leftmargin=*]
    \item[\checkmark] \textbf{Confirmed:} The reproduced MNet matches the original performance within reproducibility bounds.
    \item PROMISE reproduced at 89.0\% Dice (0.8\% from reported 89.8\%); LiTS liver reproduced at 94.3\%.
\end{itemize}
\end{minipage}
\\ \hline

\textbf{H2} & 
\begin{minipage}[t]{\linewidth}
\begin{itemize}[leftmargin=*]
    \item[\checkmark] \textbf{Confirmed:} The extensions degrade less under increased anisotropy than the baseline.
    \item On PROMISE, Spatial Gate drops 1.5 points (89.6→88.1) vs.\ 2.0 for baseline; similar preservation observed on LiTS liver.
\end{itemize}
\end{minipage}
\\ \hline

\textbf{H3} & 
\begin{minipage}[t]{\linewidth}
\begin{itemize}[leftmargin=*]
    \item[\(\triangle\)] \textbf{Partially confirmed:} Extensions improve performance under optimal spacing, with dataset-dependent gains.
    \item Spatial Gate improves PROMISE by +0.8\%; VMamba achieves best LiTS liver Dice (95.8\%) but shows higher tumor variance. 
    \item Requires further assessment via longer training on full LiTS for valid comparison.
\end{itemize}
\end{minipage}
\\ \hline

\end{tabular}
}
\vspace{4pt}
\caption{Assessment of experimental hypotheses.}
\label{tab:hypotheses}
\vspace{-25pt}
\end{table}

\subsection{Limitations and Future Work}\label{sec:limitations}
This study successfully validated the core MNet architecture and demonstrated meaningful improvements through Spatial Gate and VMamba extensions on PROMISE, achieving 89.1\% Dice with stable convergence. However, computational constraints (16GB GPU, 150 vs. 500 epochs) imposed practical limits that define clear directions for future research:

\begin{enumerate}[topsep=0pt, itemsep=0.05pt, leftmargin=*]
    \item \textit{LiTS subset evaluation:} Due to limited compute budget, LiTS experiments used 50 of 131 cases. While liver segmentation (95.8\% Dice, $\pm$0.3\%) validated cross-modality robustness, tumor results suggest that full-dataset training with extended schedules could unlock the +1-2\% headroom observed in ablations (Table~\ref{tab:ablation}, 63.8\% tumor Dice under favourable conditions).
    
    \item \textit{Convergence validation at 150 epochs:} Our loss curves and Dice score (Appendix \ref{app:A}) demonstrates Dice plateau by epoch 100-120, with 150-epoch baselines achieving 89.1\%, only 0.7\% below the 500-epoch original. This validates sufficiency for architectural comparisons, though longer schedules is likely required for LiTs which may reveal subtler benefits on irregular anatomies such as sparse tumors.
    
    \item \textit{Statistical depth via 3-fold CV:} Three-fold cross-validation balances rigor with feasibility under our constraints. The consistent PROMISE results (88.0–89.3\% across variants, $\pm$0.8–1.3\% std) provide reliable architectural insights. However, for the LiTS dataset, the extreme variance (2.8-13.6) in the tumor segmentation task is likely a result of imbalanced folds, where tumor sparsity negatively impacts outcomes. Stratified k-fold cross-validation could help mitigate this issue, and adopting a five-fold scheme would further strengthen generalization claims for clinical deployment.

    
    \item \textit{Advanced VMamba variants:} Bidirectional z-scanning (+0.2\% Dice, +40\% time) and Full 3D VMamba (OOM at 128³ patches) were implemented but remain under-explored. With 40-80GB GPUs or optimized SSM kernels, these could address datasets with complex 3D morphology (e.g., tumors, multi-organ segmentation, vascular structures).
\end{enumerate}

\textbf{Outlook.} 
Our contributions consisted of validated reproduction, learned fusion mechanisms, and efficient z-axis modelling. These establish a robust foundation for anisotropic medical imaging. The demonstrated potential stability improvements and consistent cross-modality performance (MRI/CT) confirm architectural soundness. Future work with full computational resources will likely amplify these benefits, particularly for challenging multi-class tasks and emerging state-space architectures that combine efficiency with global context modelling.

\raggedbottom 





\bibliographystyle{unsrt}  
\bibliography{references} 

\raggedbottom


\appendix



\section{Training Validation Curves}\label{app:A}

This appendix presents the complete training and validation curves for all experiments conducted in this study. The curves provide insight into convergence behavior, training stability, and the effectiveness of our architectural modifications.

\begin{figure*}[!t]
  \centering

  \begin{subfigure}{0.29\linewidth}
    \centering
    \includegraphics[width=\linewidth]{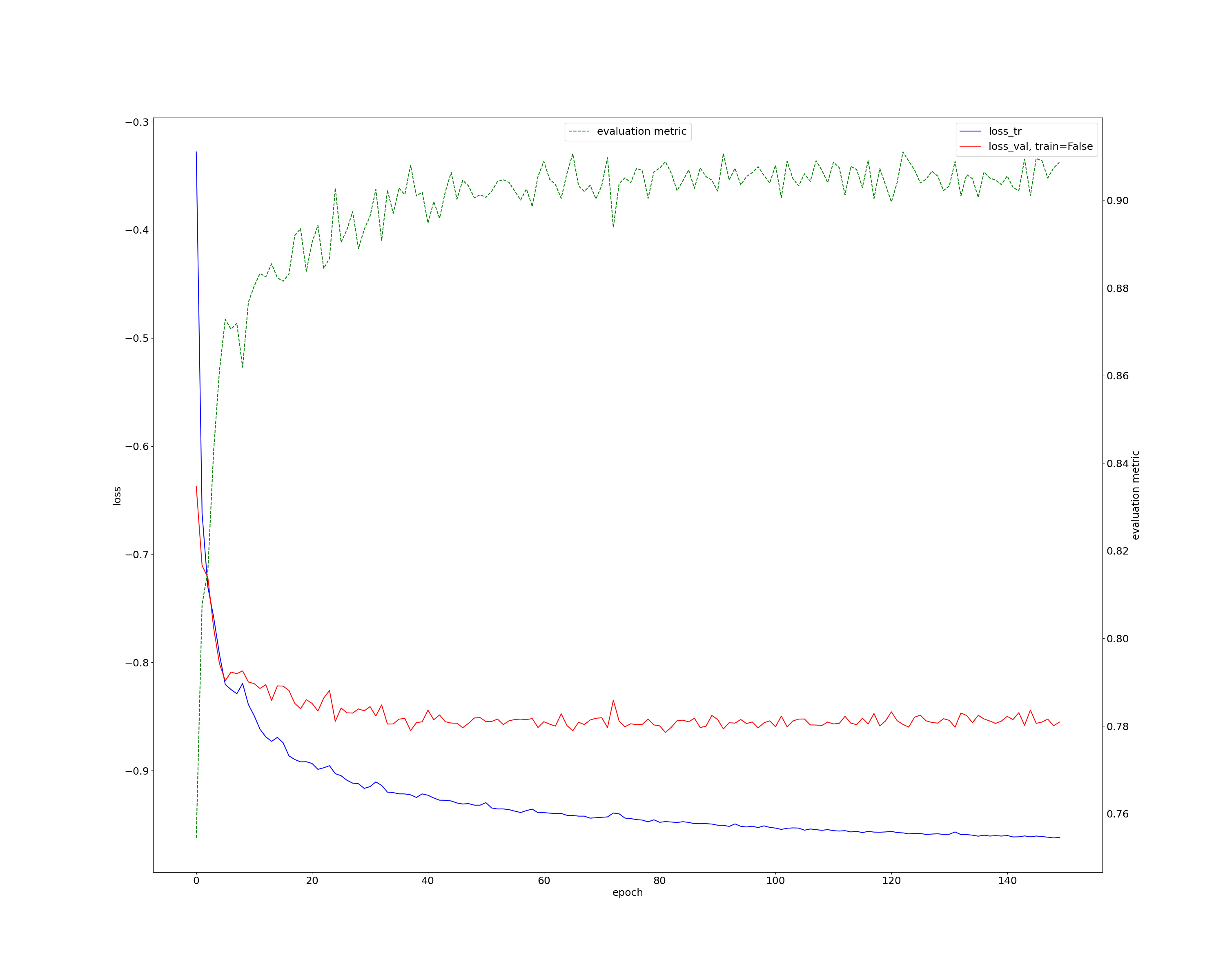}
    \caption*{\small Official — Fold 1}
  \end{subfigure}
  \hfill
  \begin{subfigure}{0.29\linewidth}
    \centering
    \includegraphics[width=\linewidth]{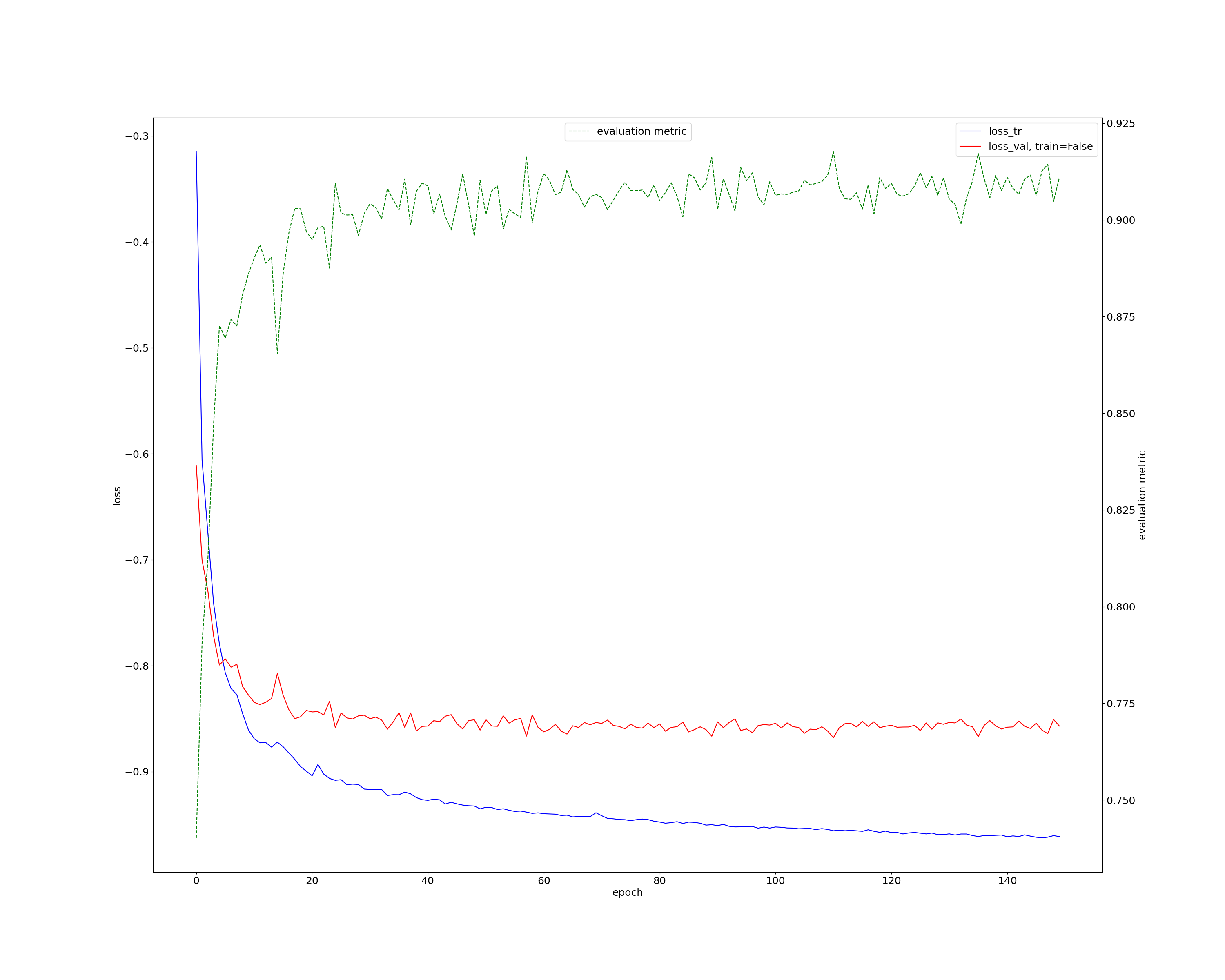}
    \caption*{\small Official — Fold 2}
  \end{subfigure}
  \hfill
  \begin{subfigure}{0.29\linewidth}
    \centering
    \includegraphics[width=\linewidth]{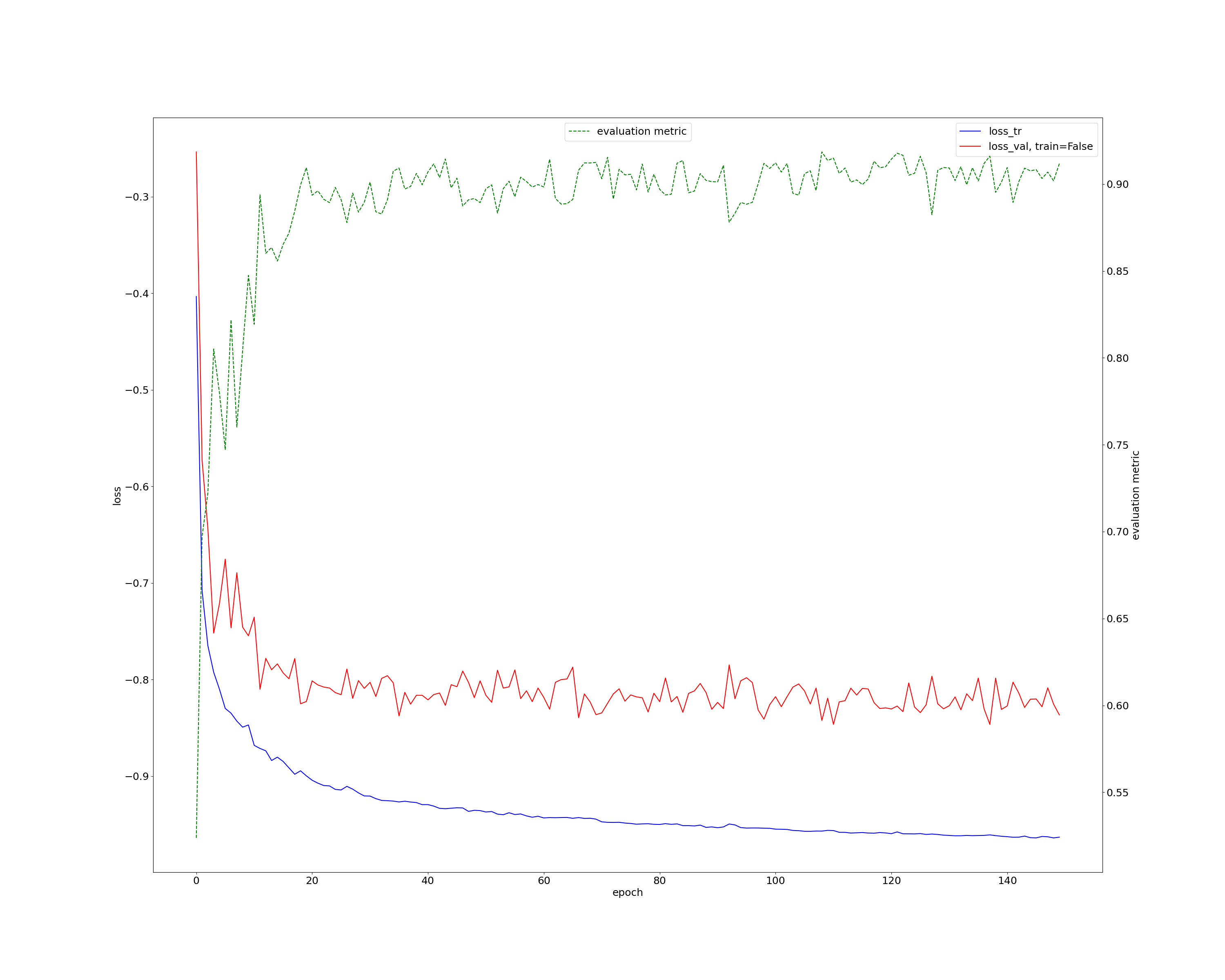}
    \caption*{\small Official — Fold 3}
  \end{subfigure}
  \\[3pt]

  \begin{subfigure}{0.29\linewidth}
    \centering
    \includegraphics[width=\linewidth]{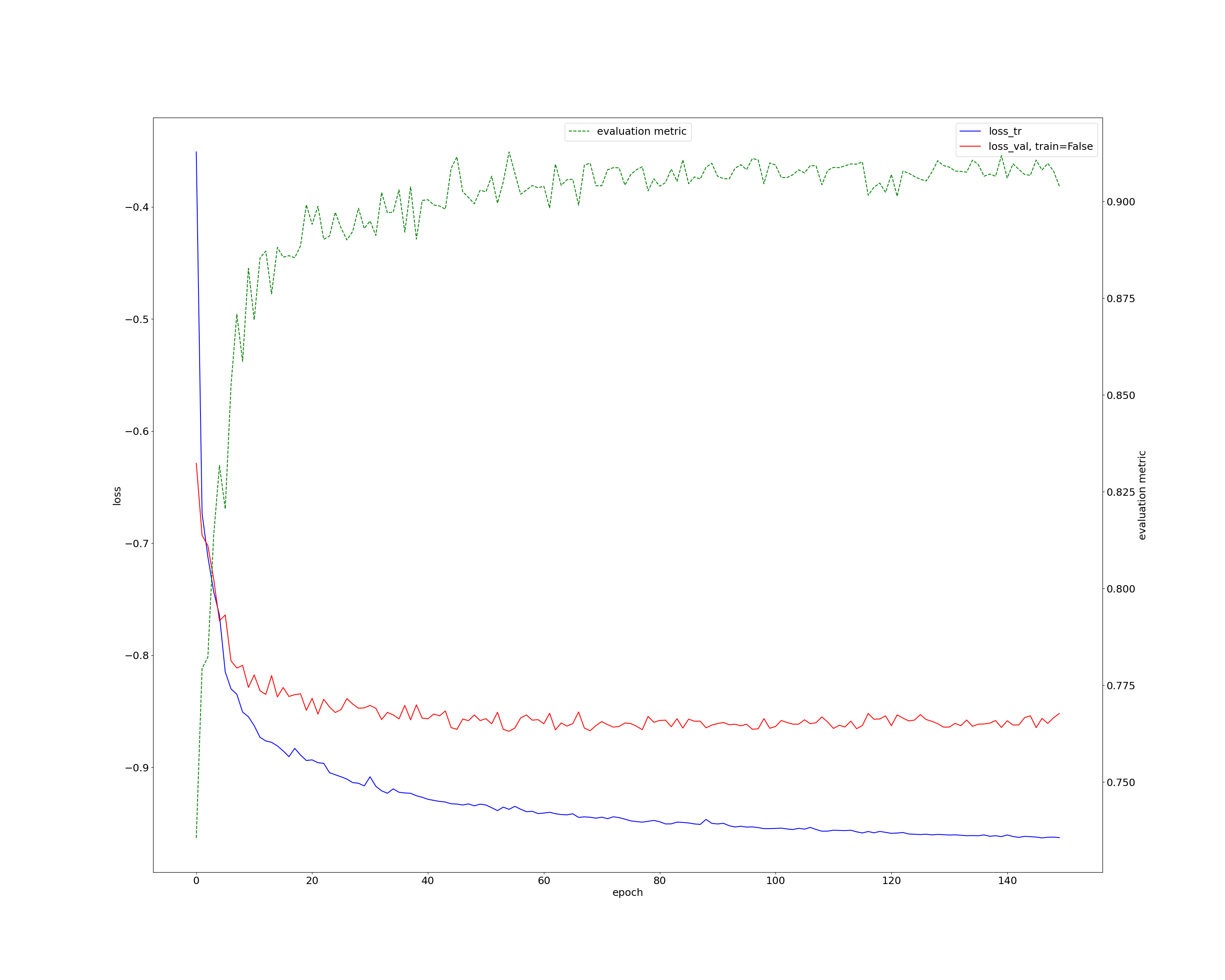}
    \caption*{\small Reproduction — Fold 1}
  \end{subfigure}
  \hfill
  \begin{subfigure}{0.29\linewidth}
    \centering
    \includegraphics[width=\linewidth]{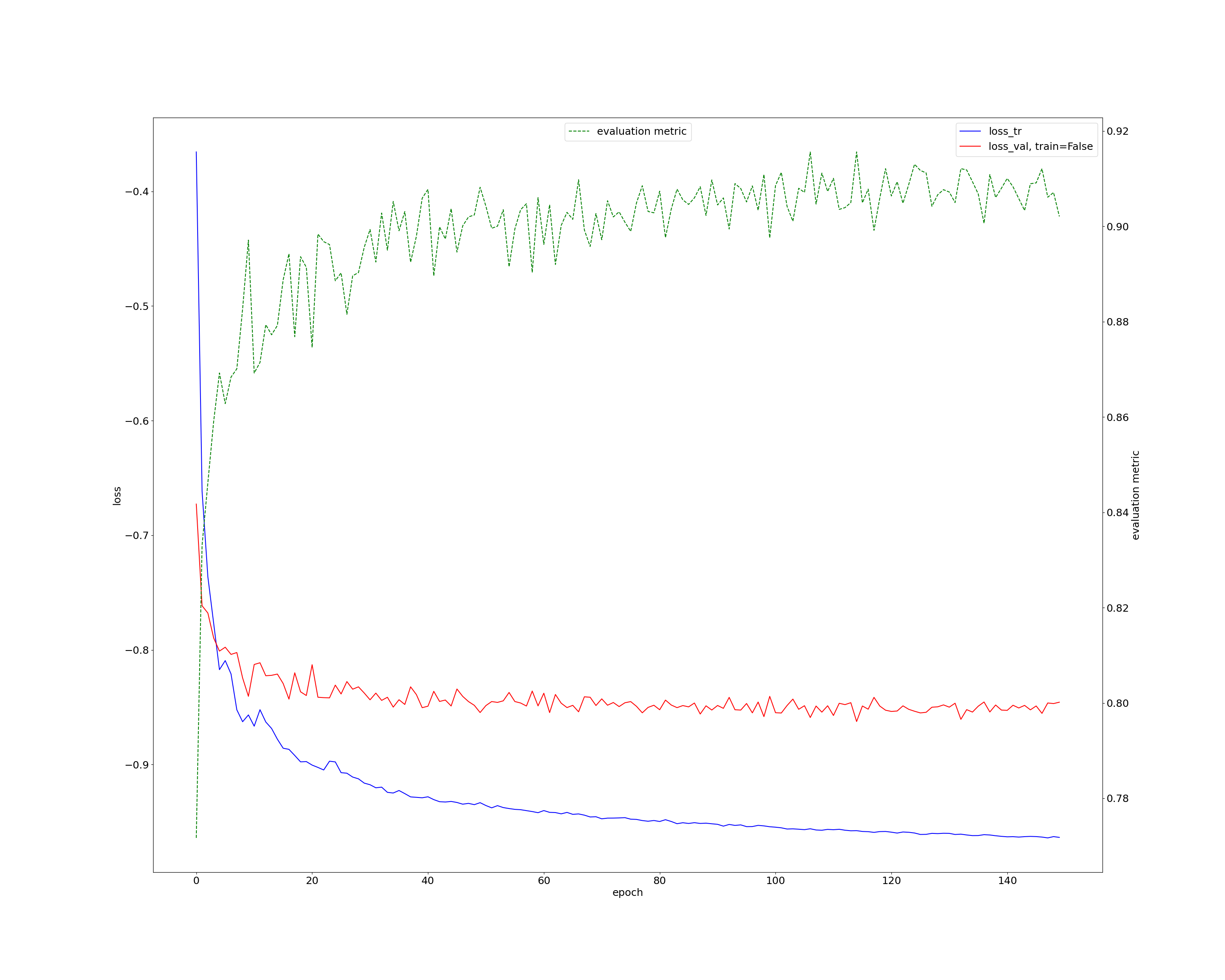}
    \caption*{\small Reproduction — Fold 2}
  \end{subfigure}
  \hfill
  \begin{subfigure}{0.29\linewidth}
    \centering
    \includegraphics[width=\linewidth]{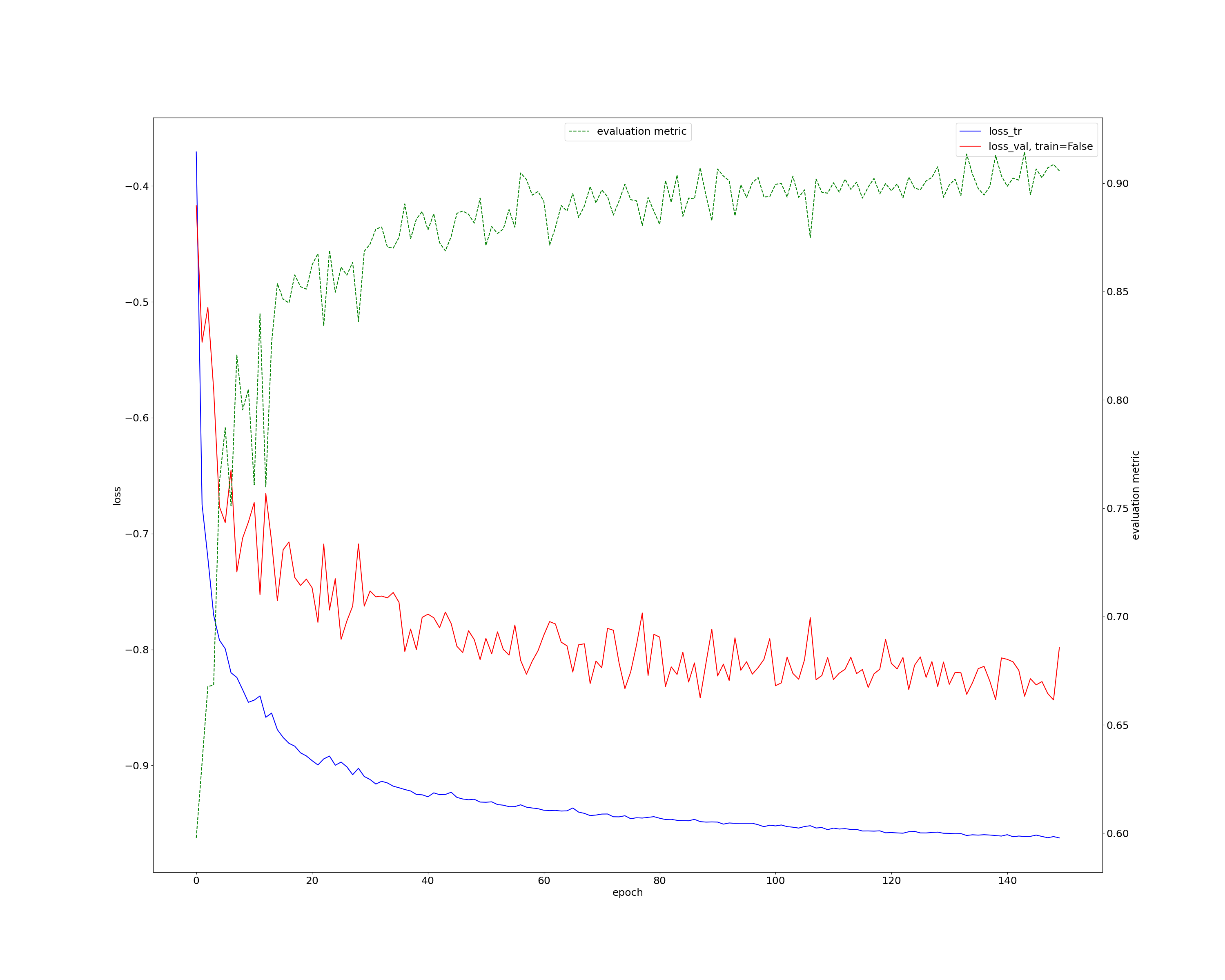}
    \caption*{\small Reproduction — Fold 3}
  \end{subfigure}
  \\[3pt]

  \begin{subfigure}{0.29\linewidth}
    \centering
    \includegraphics[width=\linewidth]{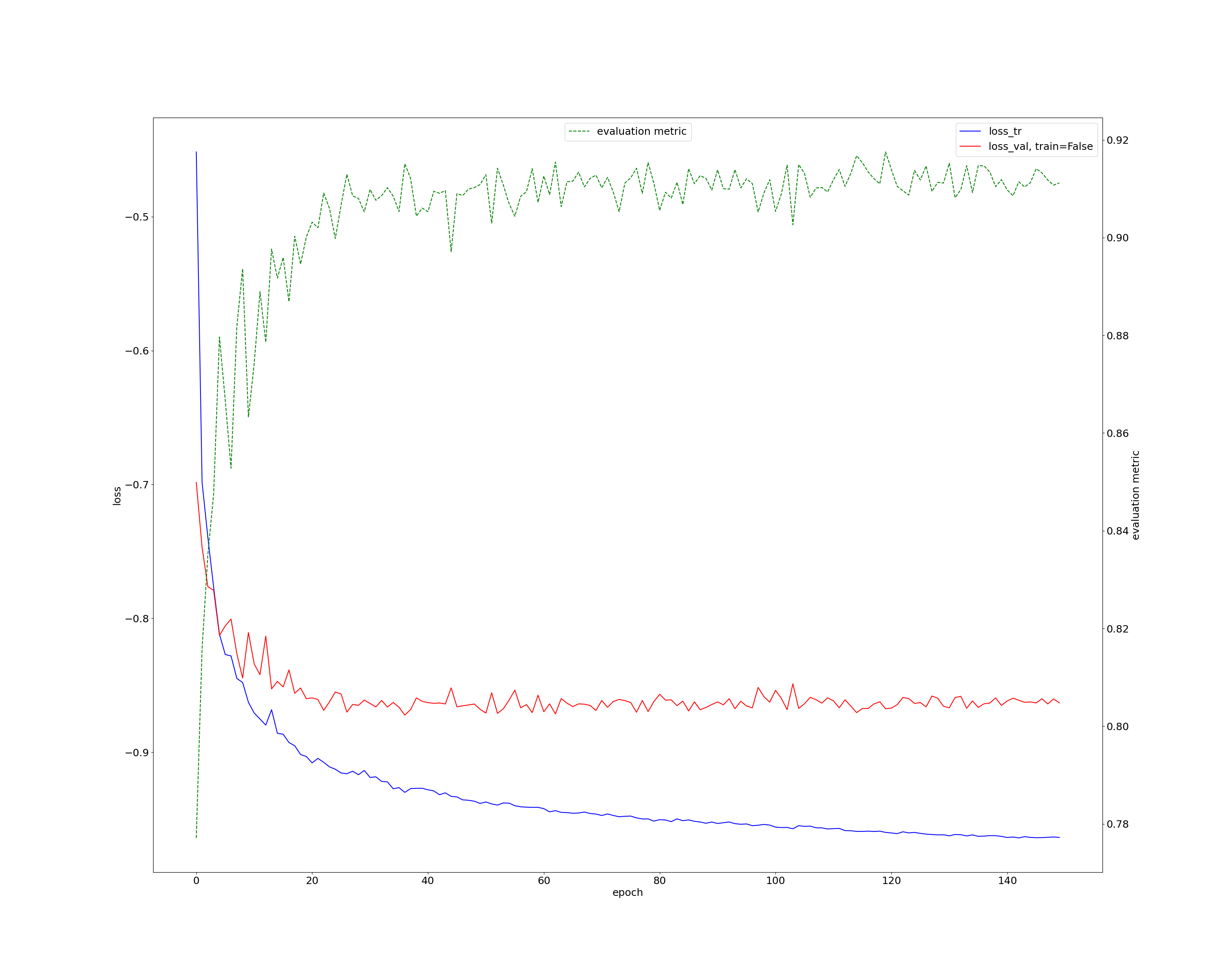}
    \caption*{\small Spatial Gate — Fold 1}
  \end{subfigure}
  \hfill
  \begin{subfigure}{0.29\linewidth}
    \centering
    \includegraphics[width=\linewidth]{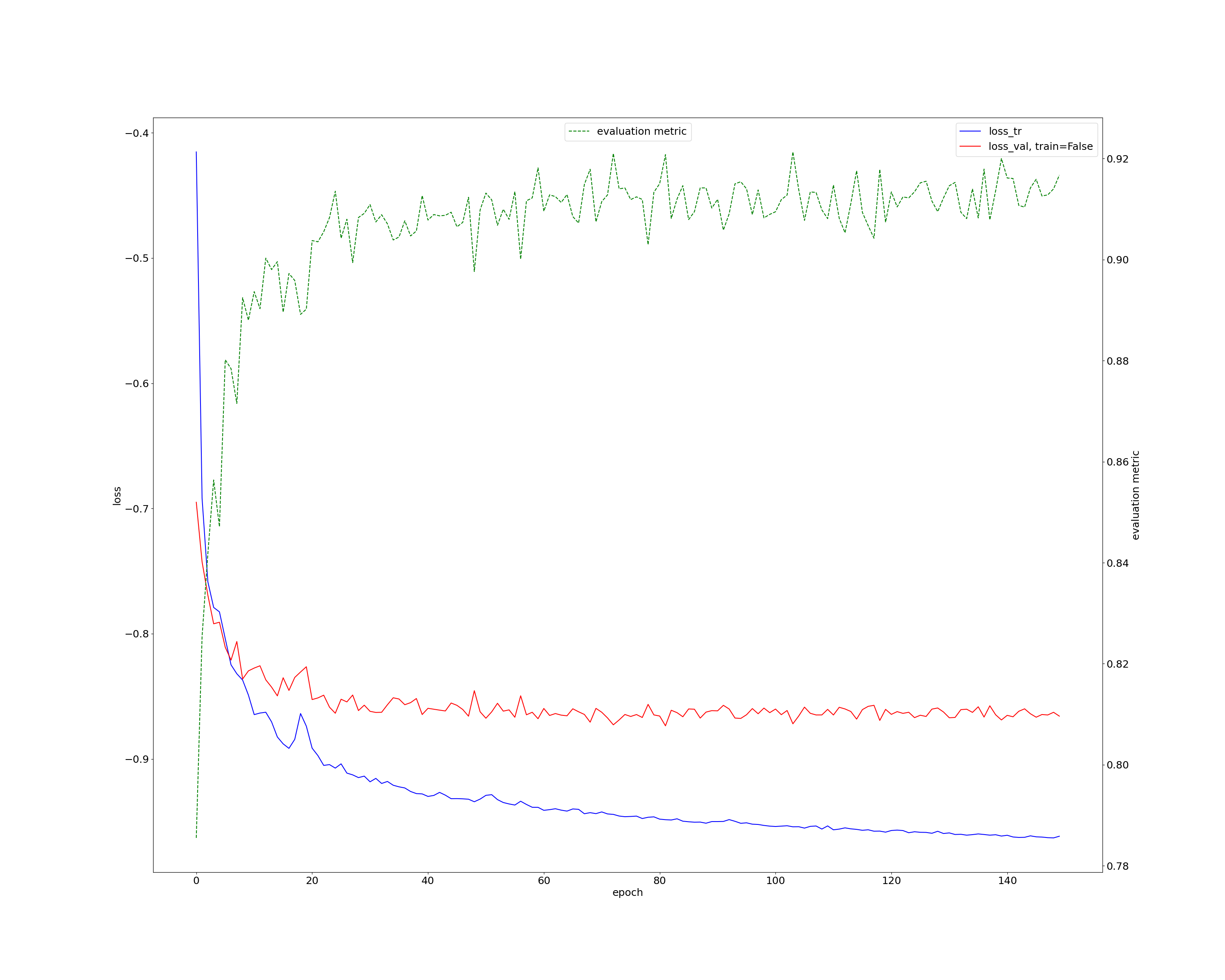}
    \caption*{\small Spatial Gate — Fold 2}
  \end{subfigure}
  \hfill
  \begin{subfigure}{0.29\linewidth}
    \centering
    \includegraphics[width=\linewidth]{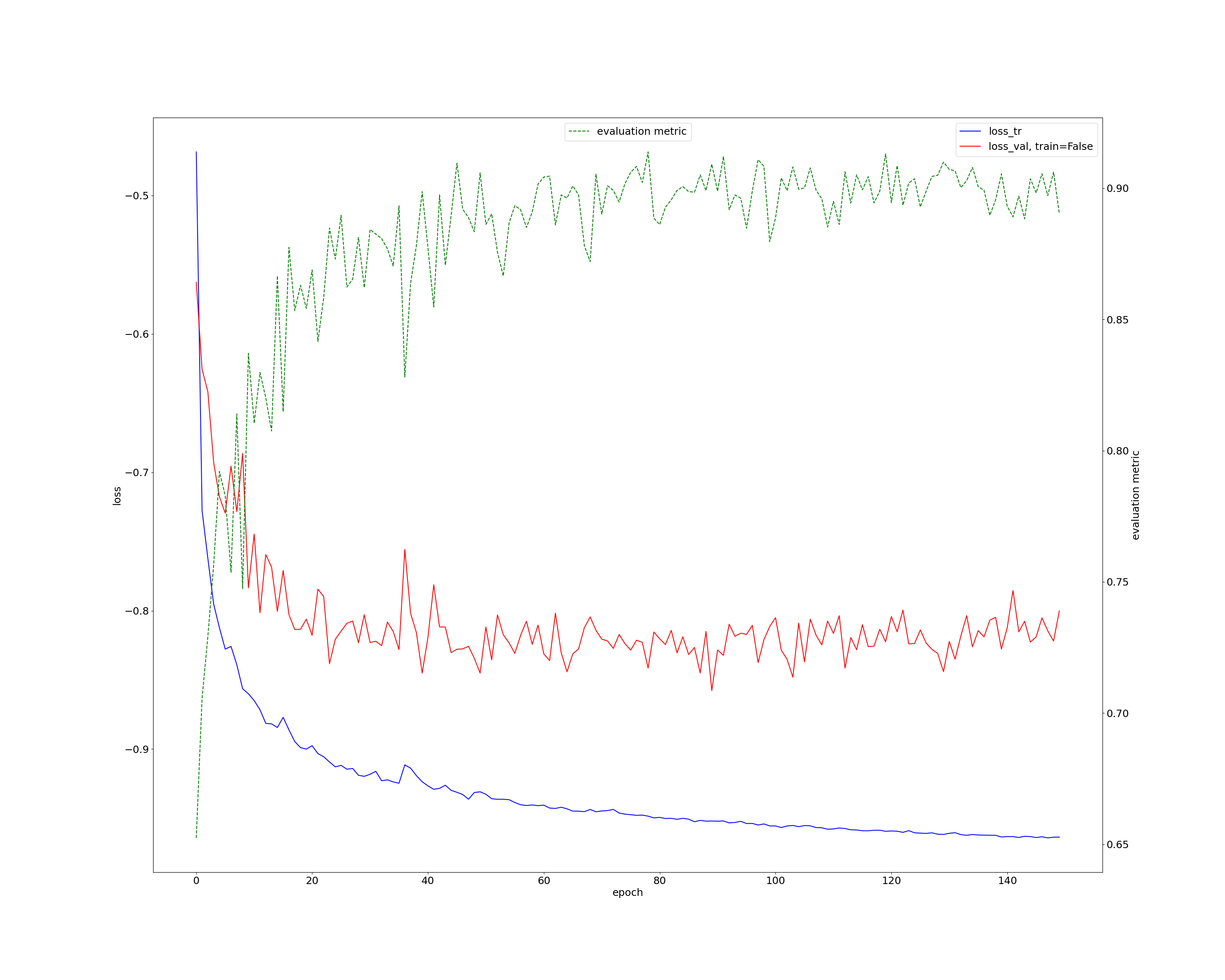}
    \caption*{\small Spatial Gate — Fold 3}
  \end{subfigure}
  \\[3pt]

  \begin{subfigure}{0.29\linewidth}
    \centering
    \includegraphics[width=\linewidth]{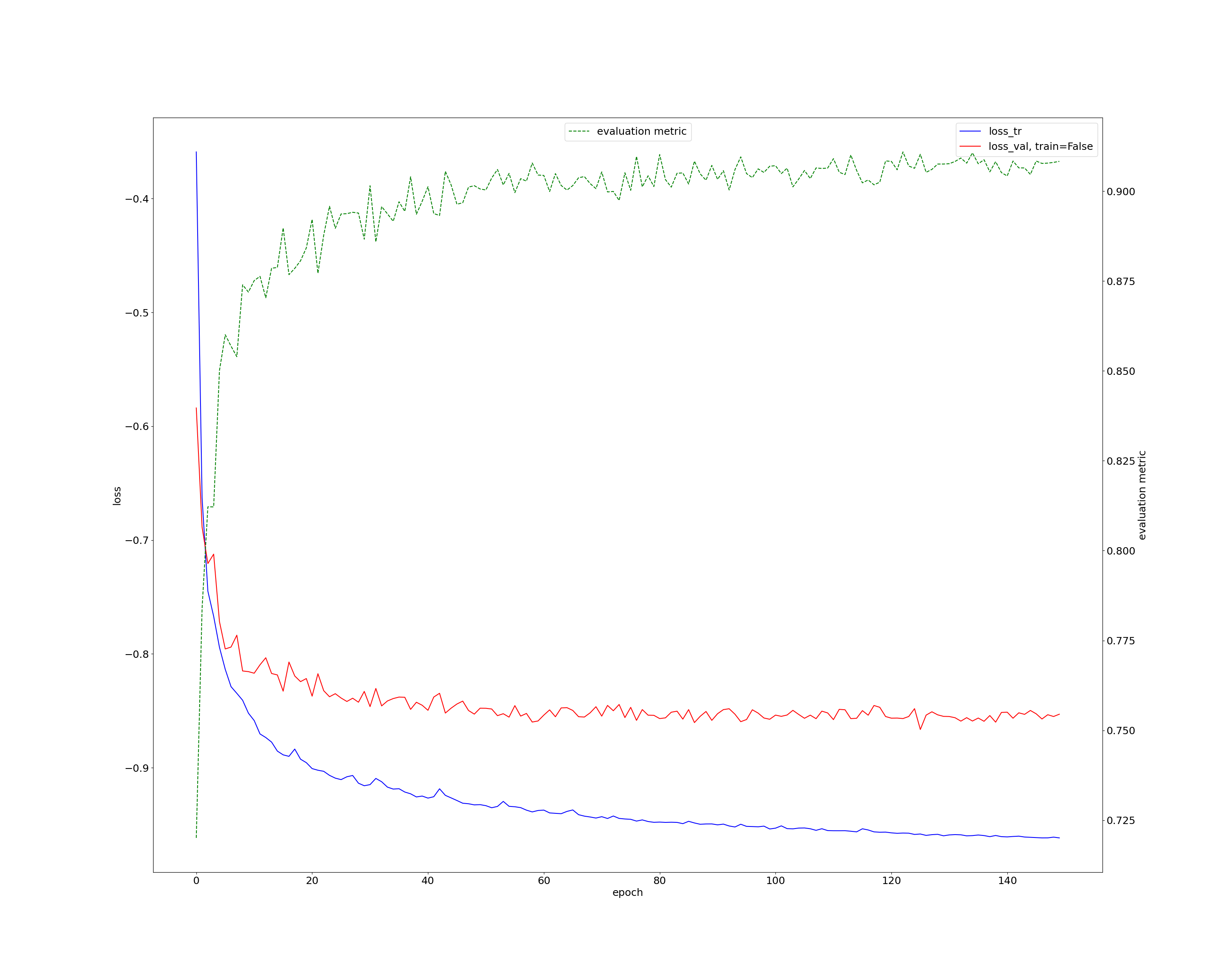}
    \caption*{\small VMamba — Fold 0}
  \end{subfigure}
  \hfill
  \begin{subfigure}{0.29\linewidth}
    \centering
    \includegraphics[width=\linewidth]{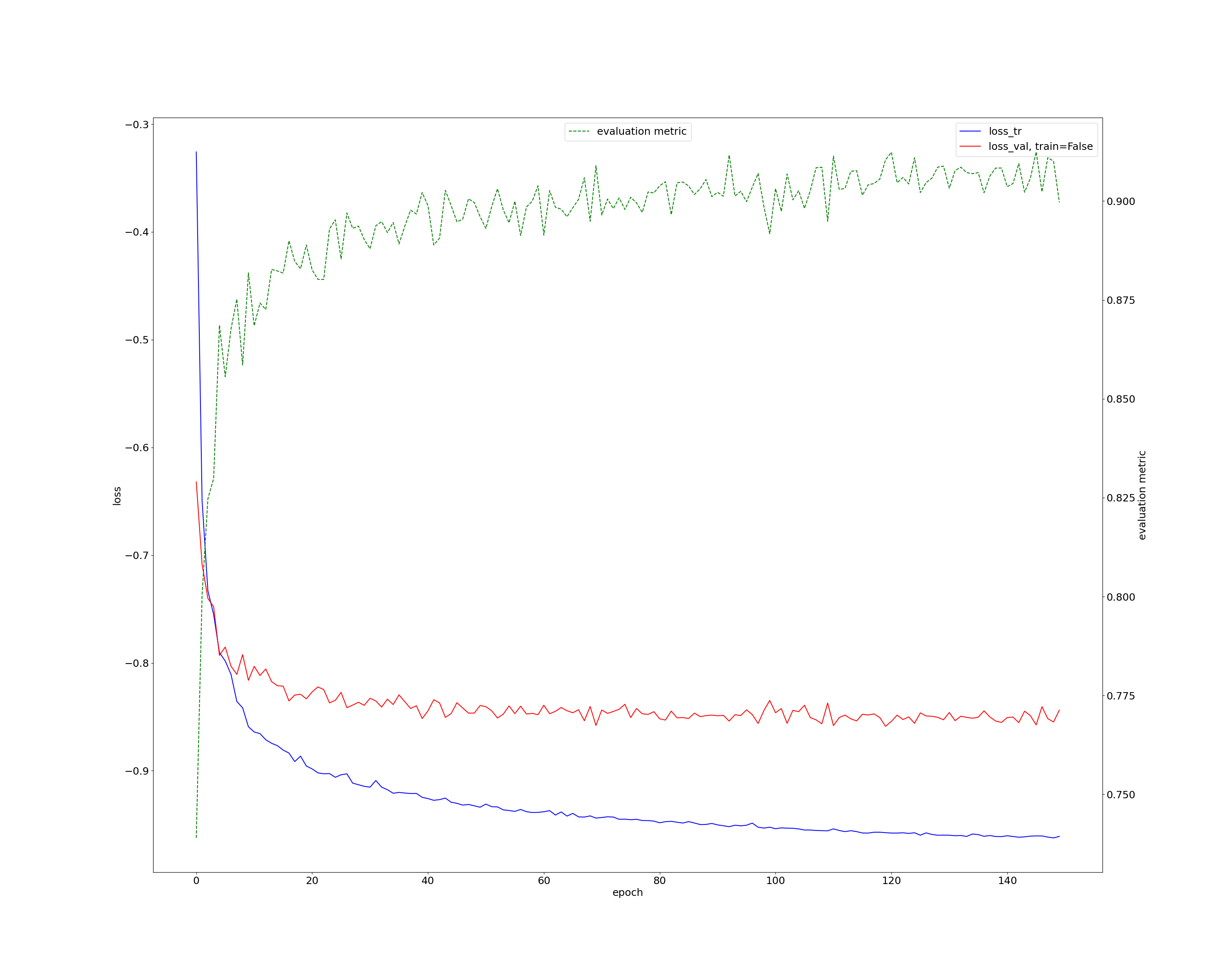}
    \caption*{\small VMamba — Fold 1}
  \end{subfigure}
  \hfill
  \begin{subfigure}{0.29\linewidth}
    \centering
    \includegraphics[width=\linewidth]{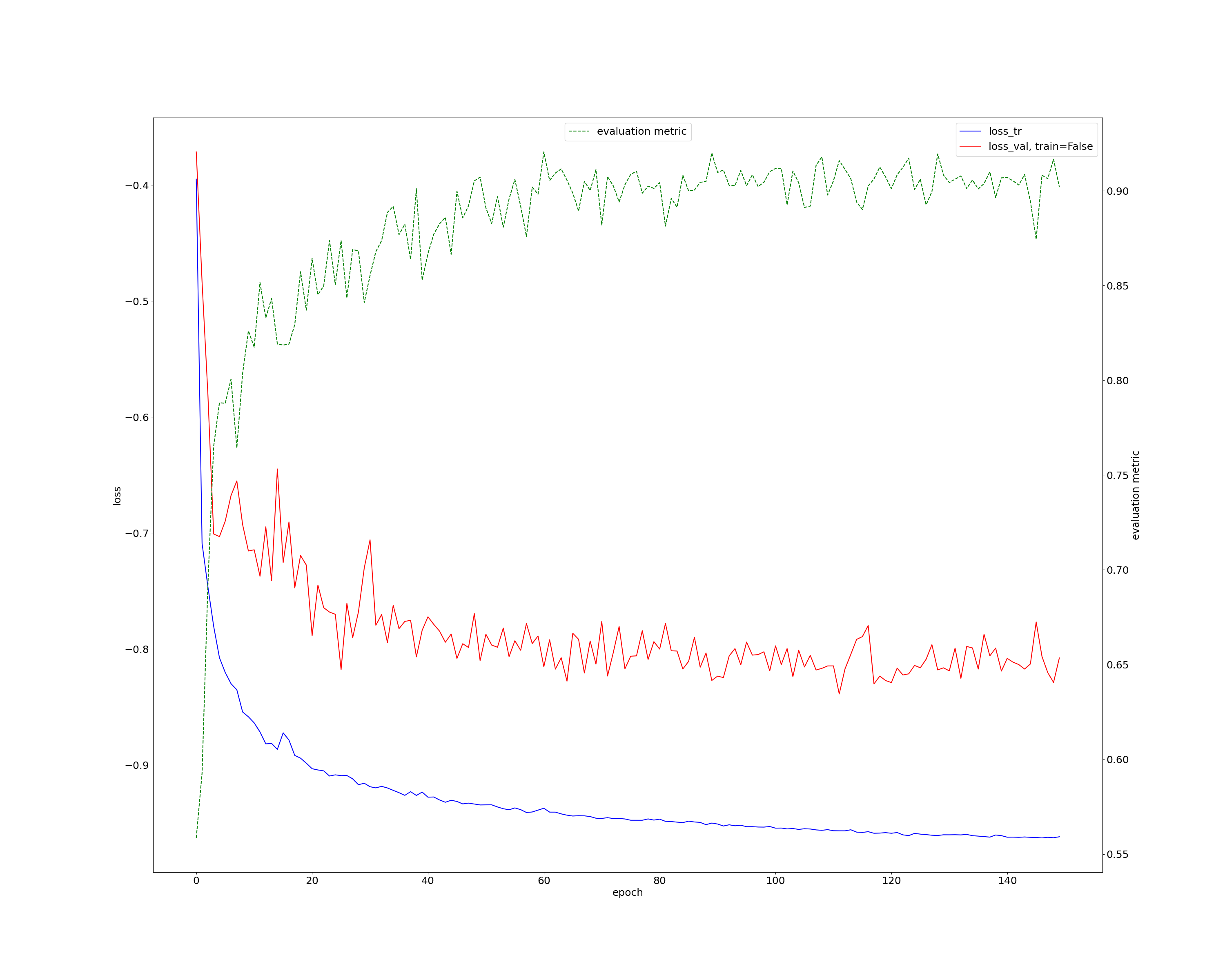}
    \caption*{\small VMamba — Fold 2}
  \end{subfigure}
  \\[3pt]

  \begin{subfigure}{0.29\linewidth}
    \centering
    \includegraphics[width=\linewidth]{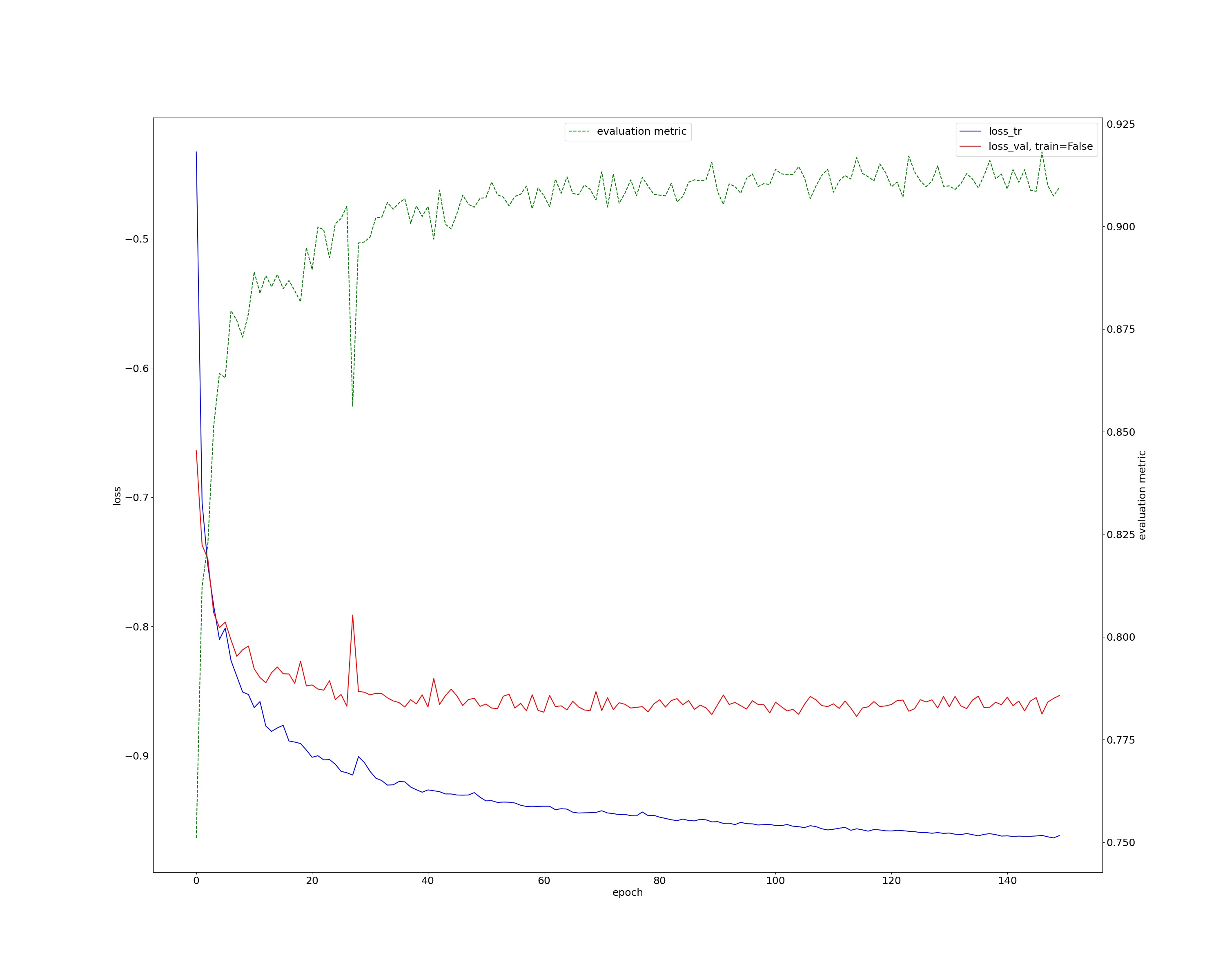}
    \caption*{\small SG + VMamba — Fold 0}
  \end{subfigure}
  \hfill
  \begin{subfigure}{0.29\linewidth}
    \centering
    \includegraphics[width=\linewidth]{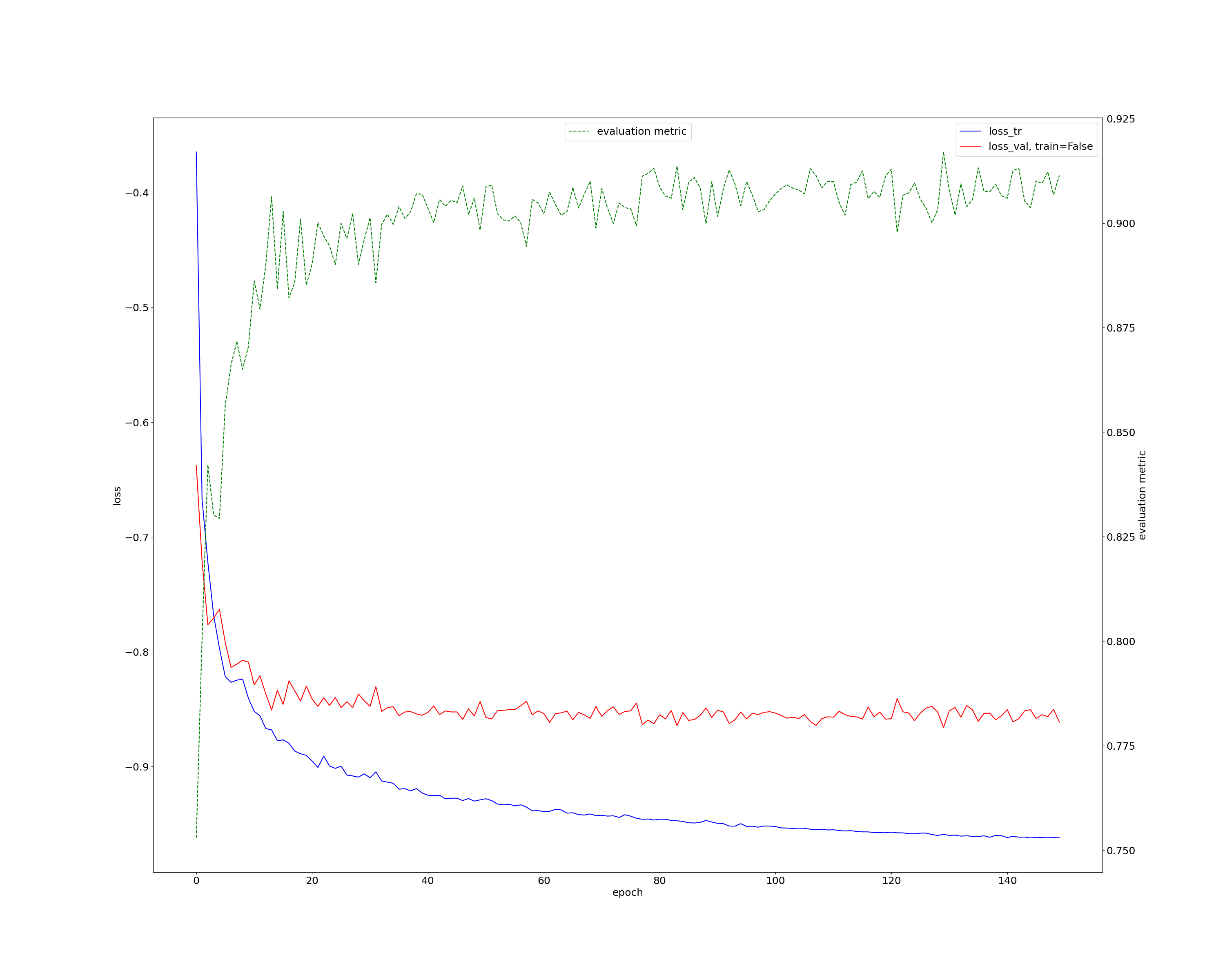}
    \caption*{\small SG + VMamba — Fold 1}
  \end{subfigure}
  \hfill
  \begin{subfigure}{0.29\linewidth}
    \centering
    \includegraphics[width=\linewidth]{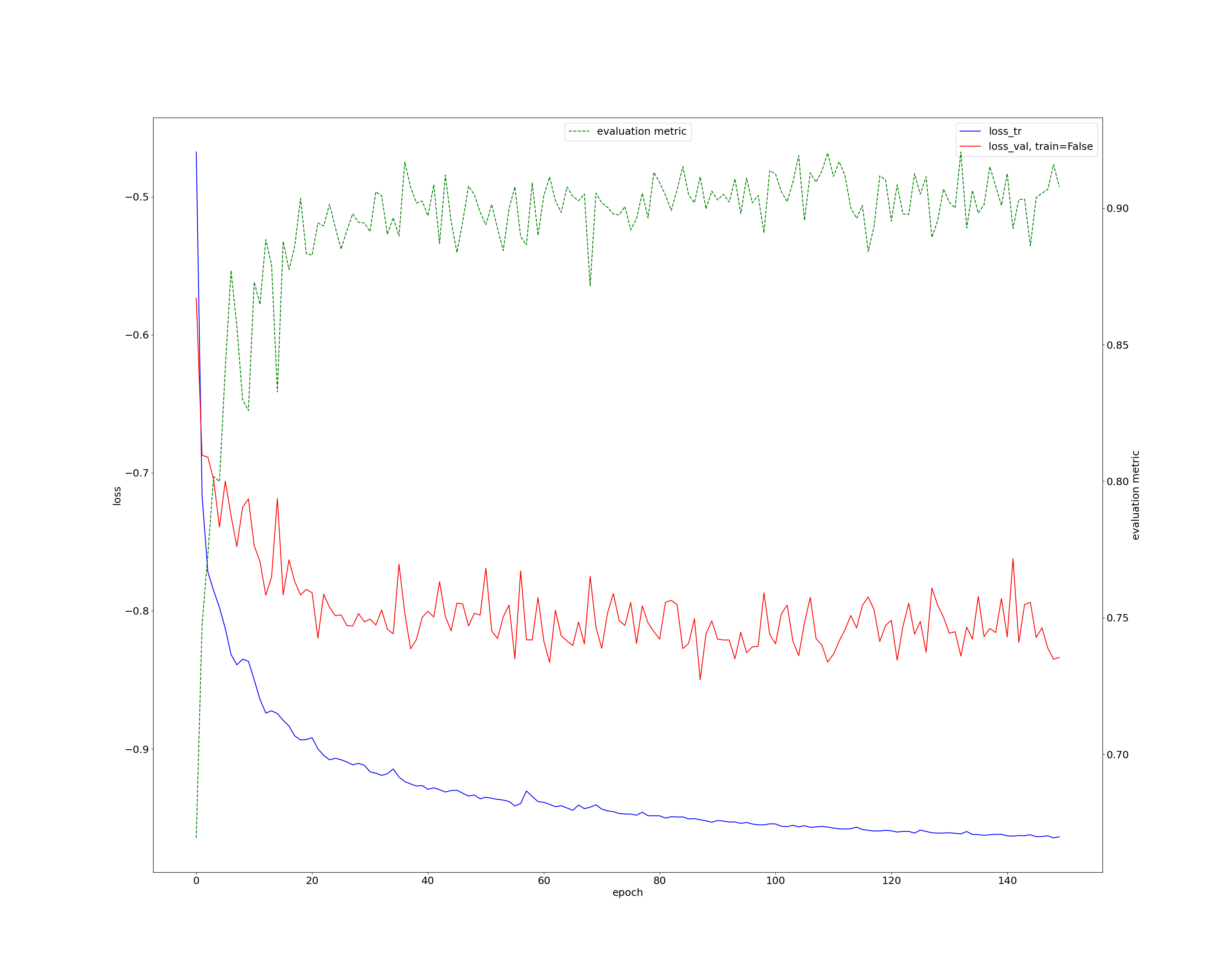}
    \caption*{\small SG + VMamba — Fold 2}
  \end{subfigure}

  \caption{PROMISE Loss for 150 epochs across 3 Folds.}
  \label{fig:promise_loss}
\end{figure*}

\textbf{PROMISE Convergence.} Figure~\ref{fig:promise_loss} presents training and validation curves across all five experimental configurations and three folds. Training shows stable convergence with smooth Dice improvements and low variance. The baseline reproduction (Rows 1–2) demonstrates consistent optimization under the nnU-Net preprocessing pipeline and Dice–cross-entropy loss. Extension variants (Rows 3–5: Spatial Gate, VMamba, and Combined SG+VMamba) maintain this stability while showing modest improvements in validation Dice. All architectural modifications preserve training dynamics without introducing optimization instabilities, validating their successful integration into the MNet framework.

\begin{figure*}[!t]
  \centering

  \begin{subfigure}{0.29\linewidth}
    \centering
    \includegraphics[width=\linewidth]{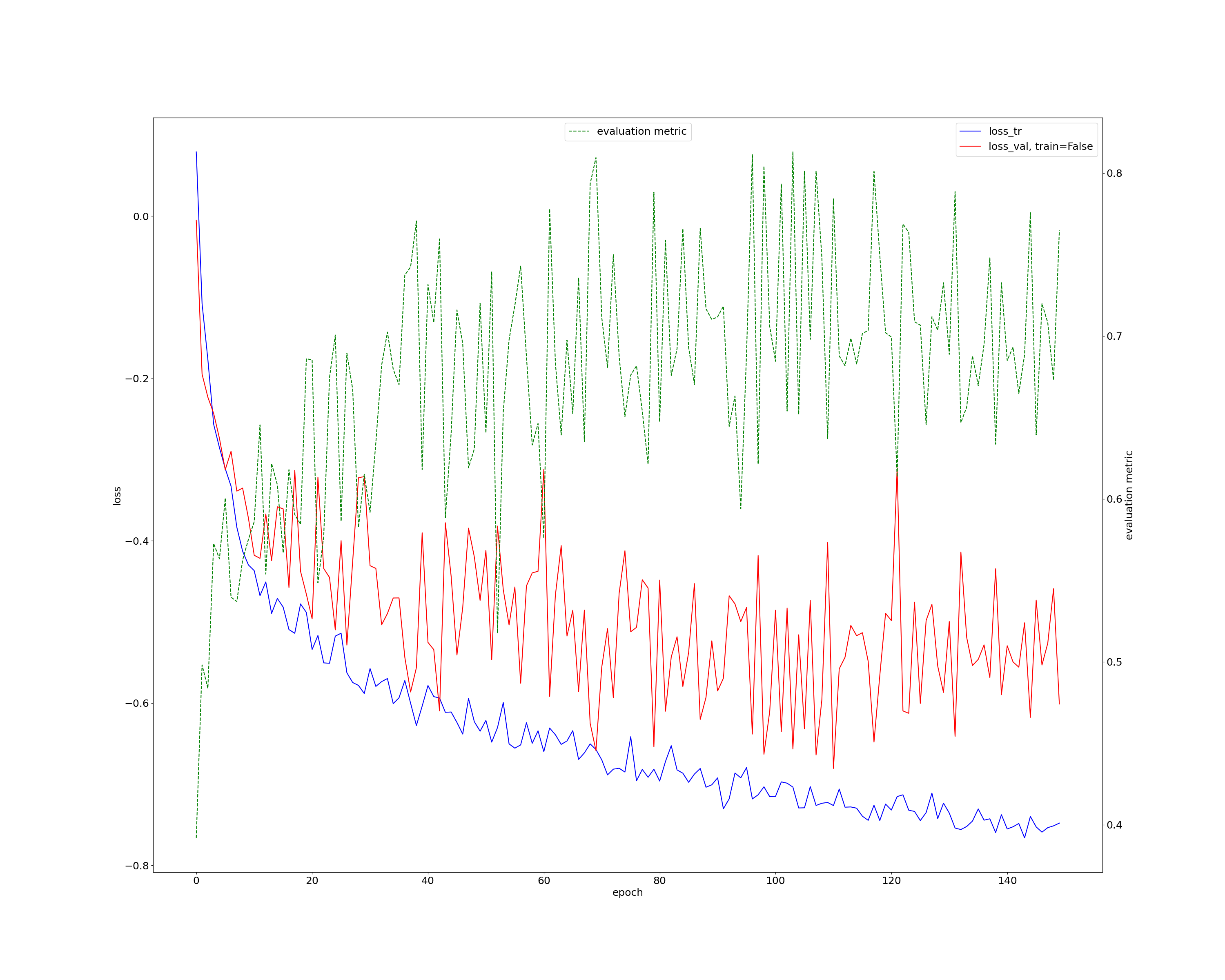}
    \caption*{\small Official — Fold 1}
  \end{subfigure}
  \hfill
  \begin{subfigure}{0.29\linewidth}
    \centering
    \includegraphics[width=\linewidth]{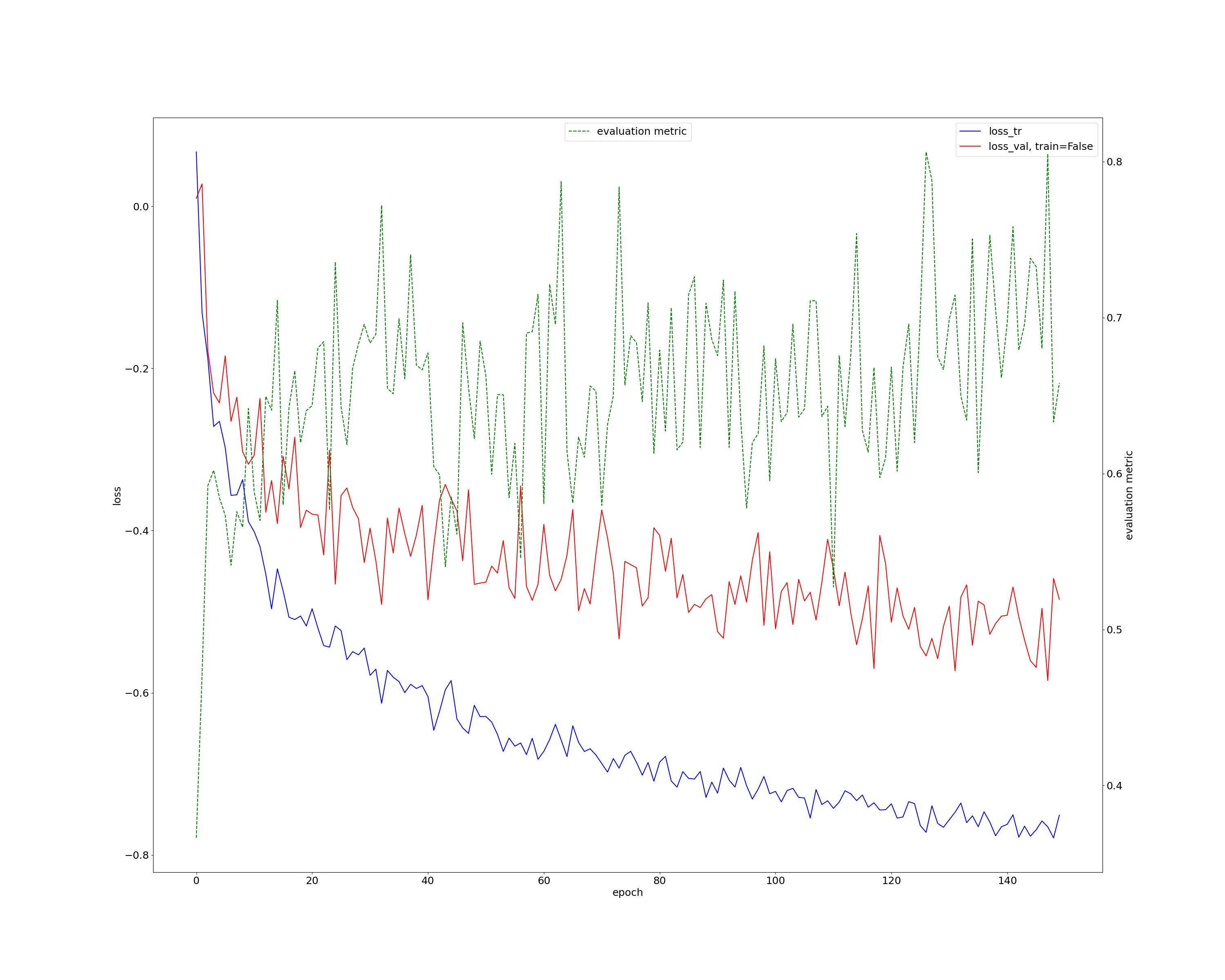}
    \caption*{\small Official — Fold 2}
  \end{subfigure}
  \hfill
  \begin{subfigure}{0.29\linewidth}
    \centering
    \includegraphics[width=\linewidth]{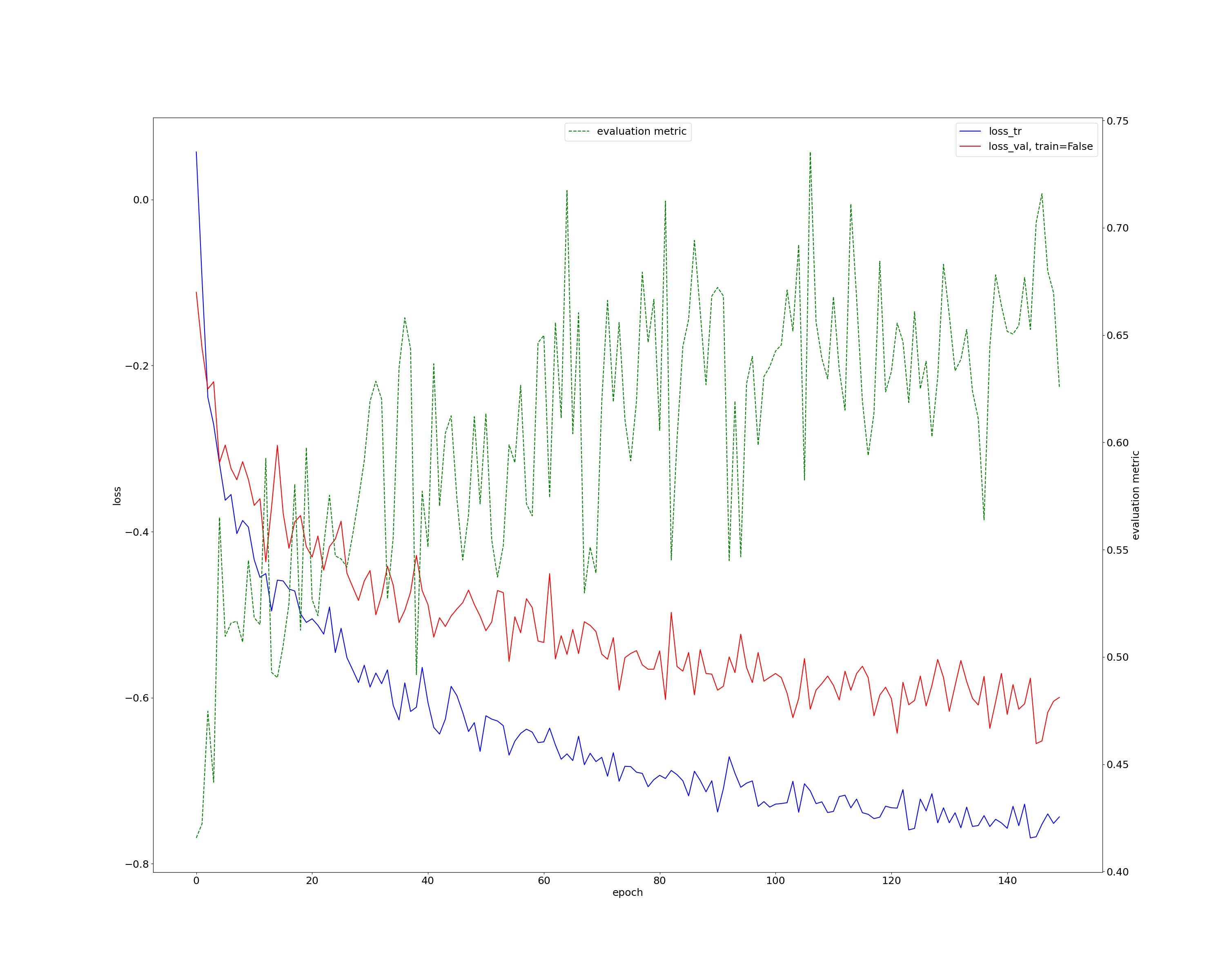}
    \caption*{\small Official — Fold 3}
  \end{subfigure}
  \\[3pt]

  \begin{subfigure}{0.29\linewidth}
    \centering
    \includegraphics[width=\linewidth]{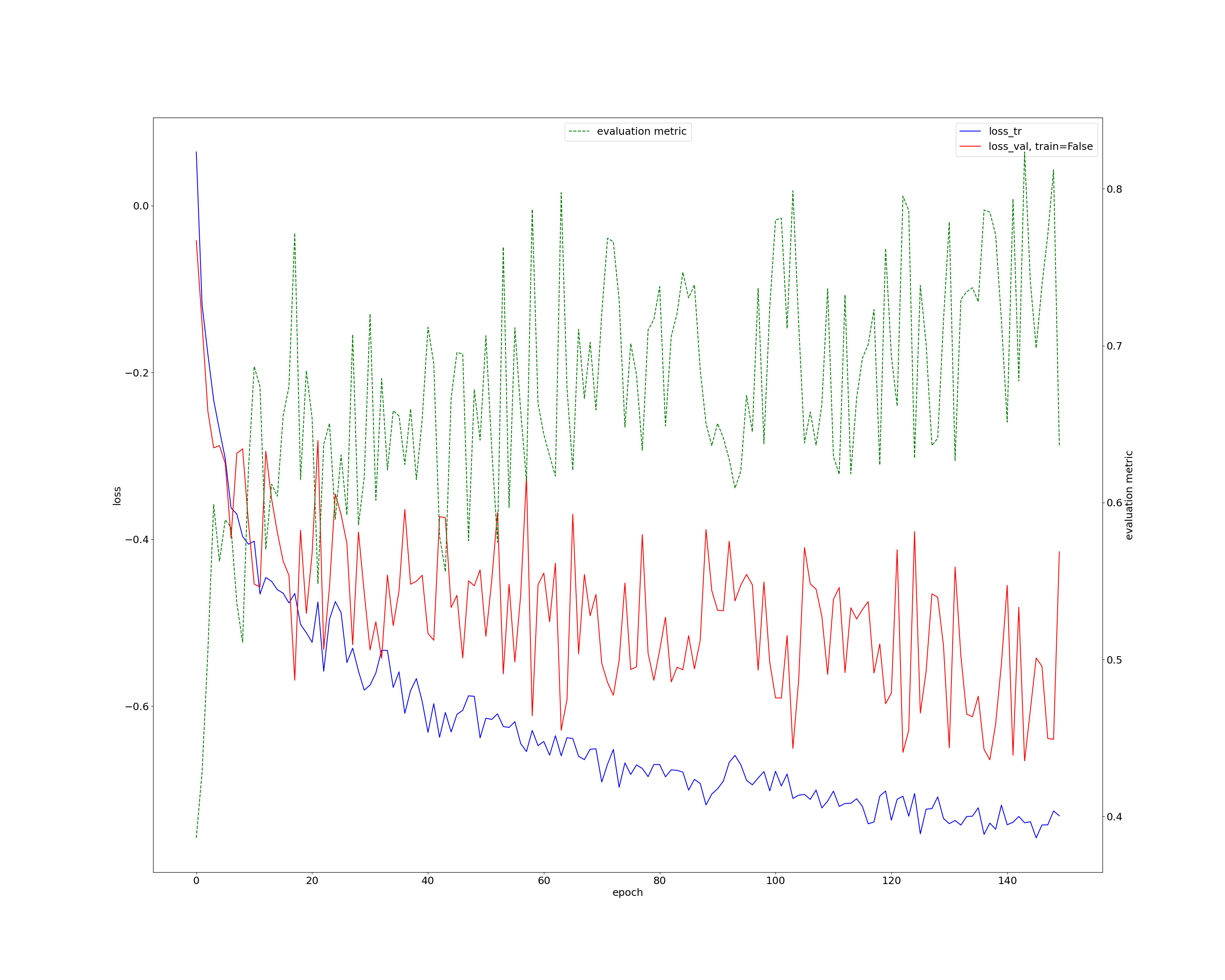}
    \caption*{\small Reproduction — Fold 1}
  \end{subfigure}
  \hfill
  \begin{subfigure}{0.29\linewidth}
    \centering
    \includegraphics[width=\linewidth]{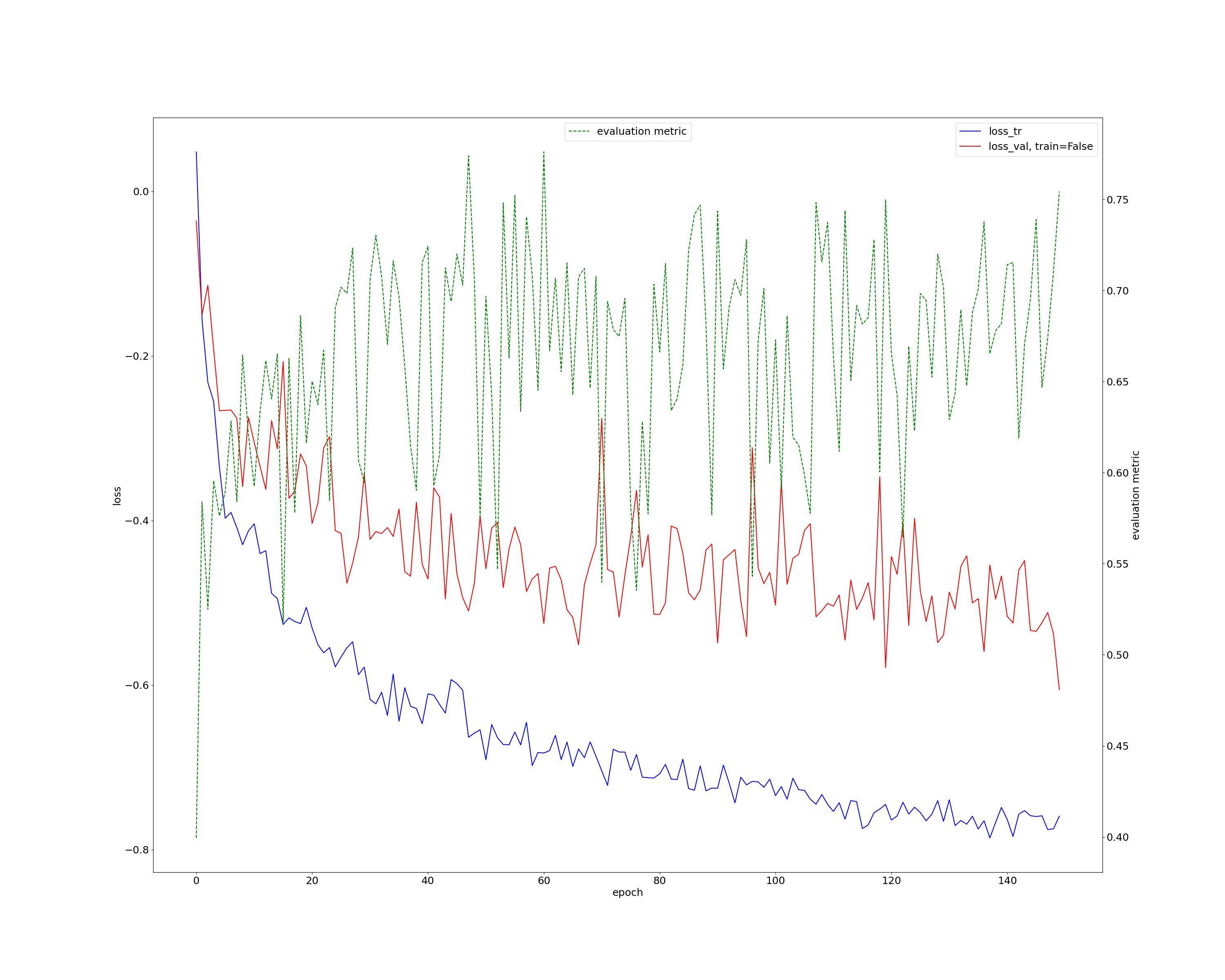}
    \caption*{\small Reproduction — Fold 2}
  \end{subfigure}
  \hfill
  \begin{subfigure}{0.29\linewidth}
    \centering
    \includegraphics[width=\linewidth]{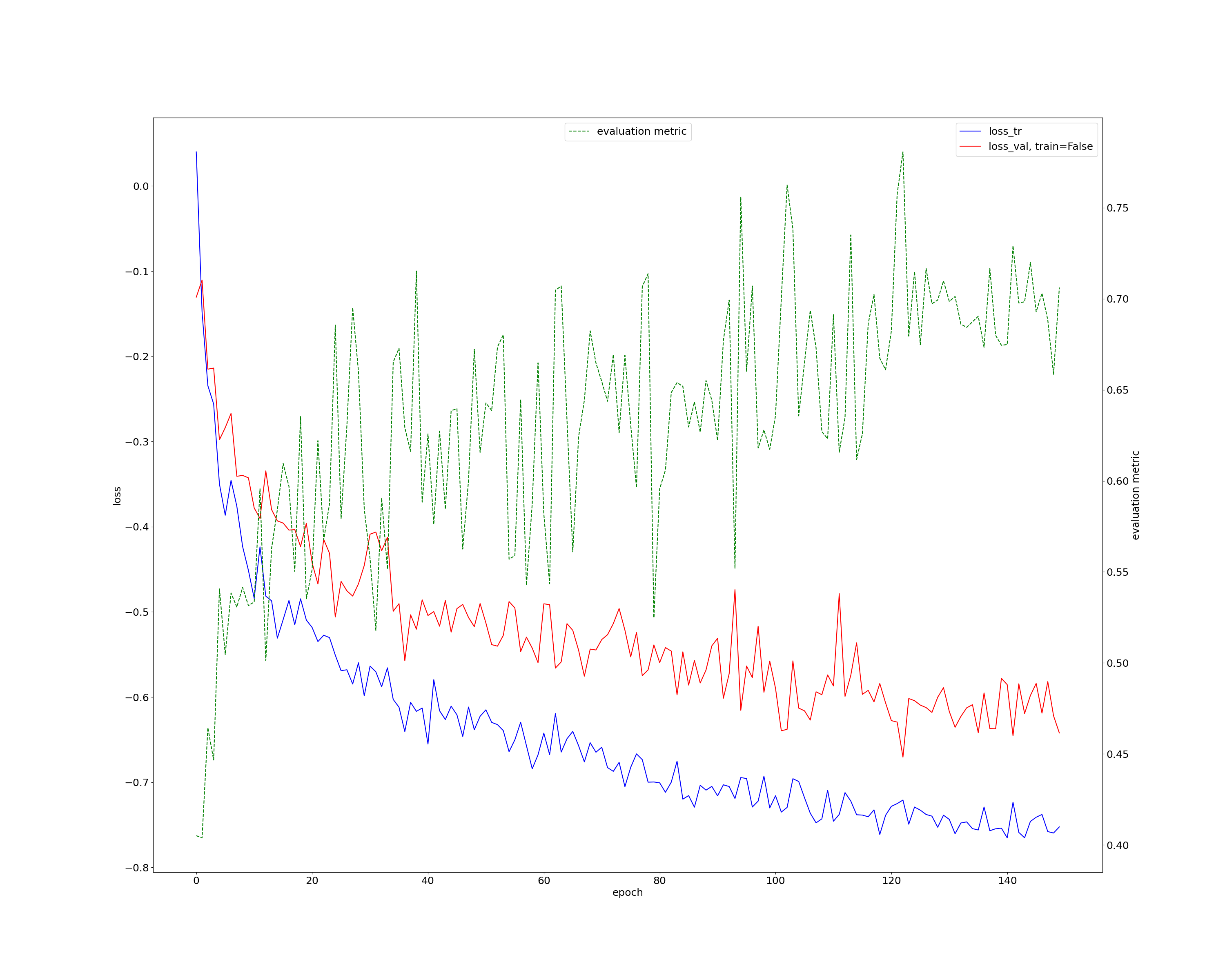}
    \caption*{\small Reproduction — Fold 3}
  \end{subfigure}
  \\[3pt]

  \begin{subfigure}{0.29\linewidth}
    \centering
    \includegraphics[width=\linewidth]{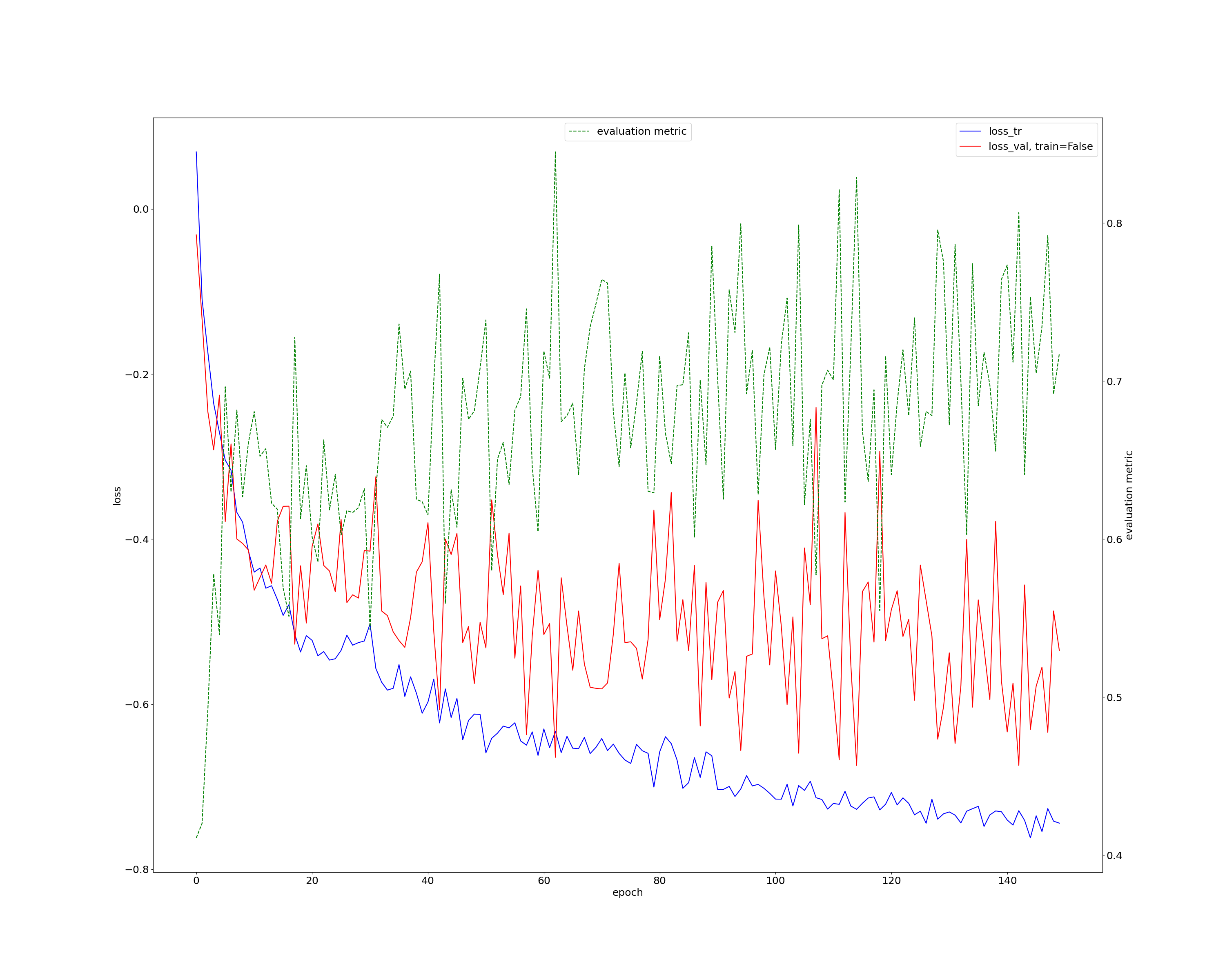}
    \caption*{\small Spatial Gate — Fold 1}
  \end{subfigure}
  \hfill
  \begin{subfigure}{0.29\linewidth}
    \centering
    \includegraphics[width=\linewidth]{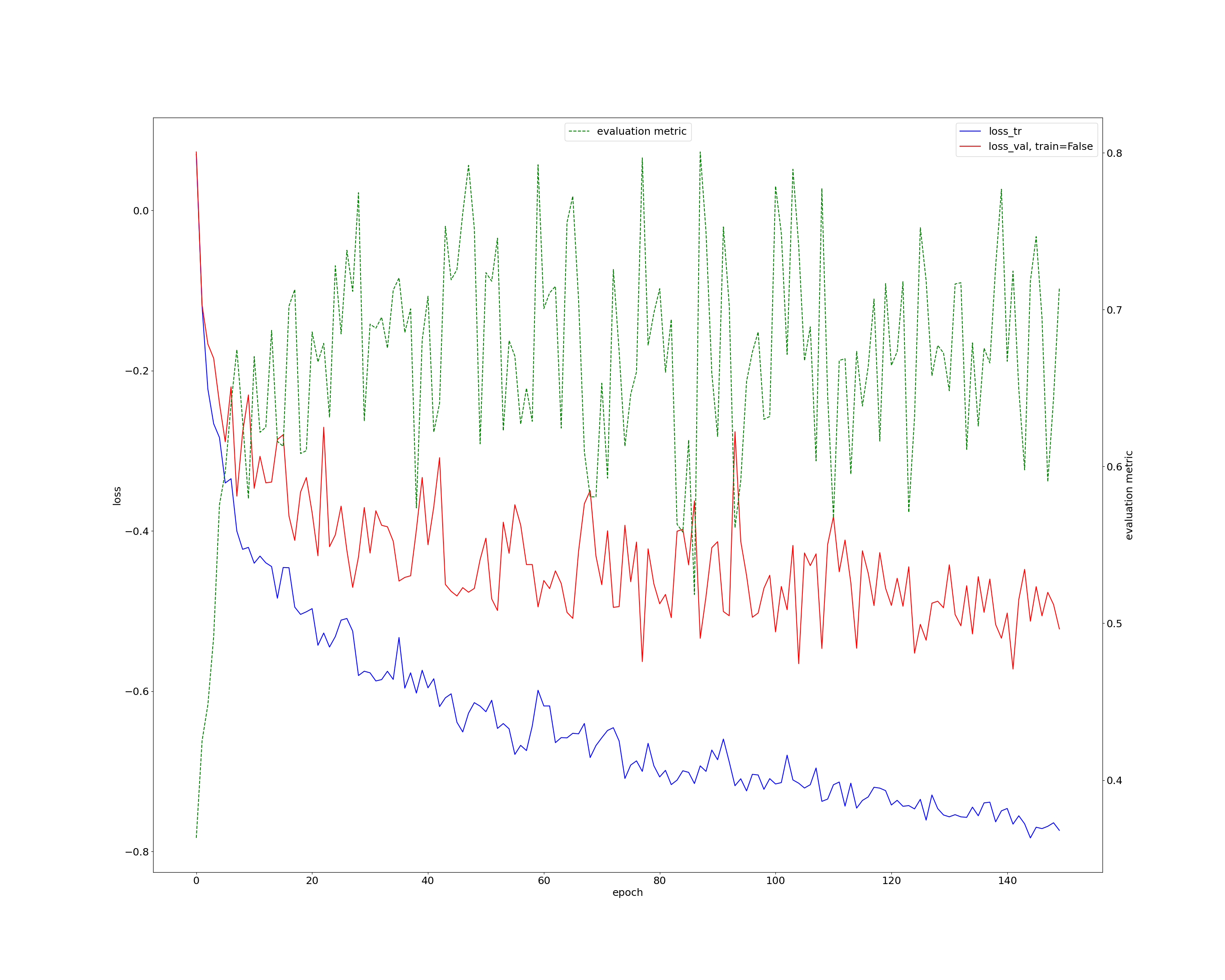}
    \caption*{\small Spatial Gate — Fold 2}
  \end{subfigure}
  \hfill
  \begin{subfigure}{0.29\linewidth}
    \centering
    \includegraphics[width=\linewidth]{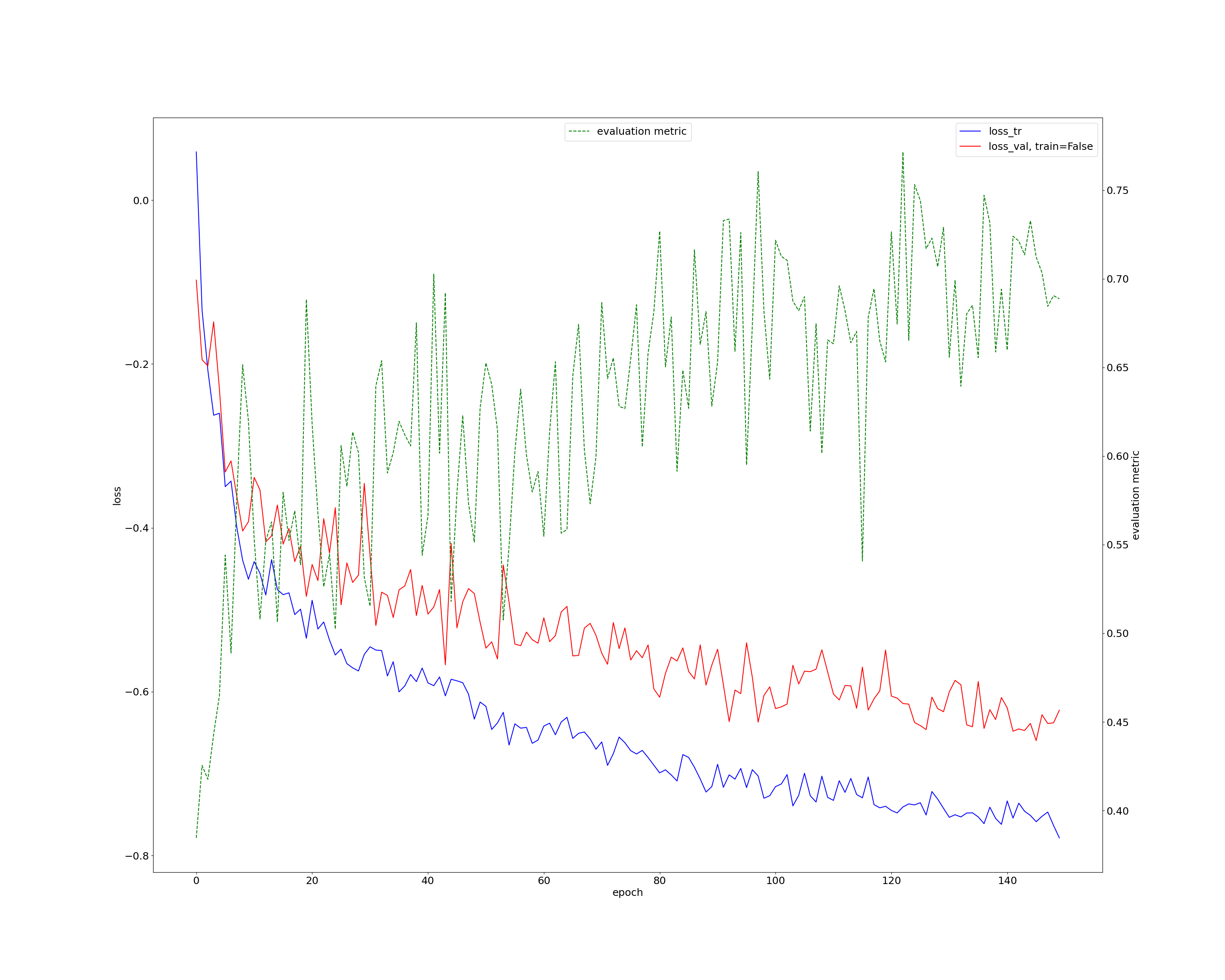}
    \caption*{\small Spatial Gate — Fold 3}
  \end{subfigure}
  \\[3pt]

  \begin{subfigure}{0.29\linewidth}
    \centering
    \includegraphics[width=\linewidth]{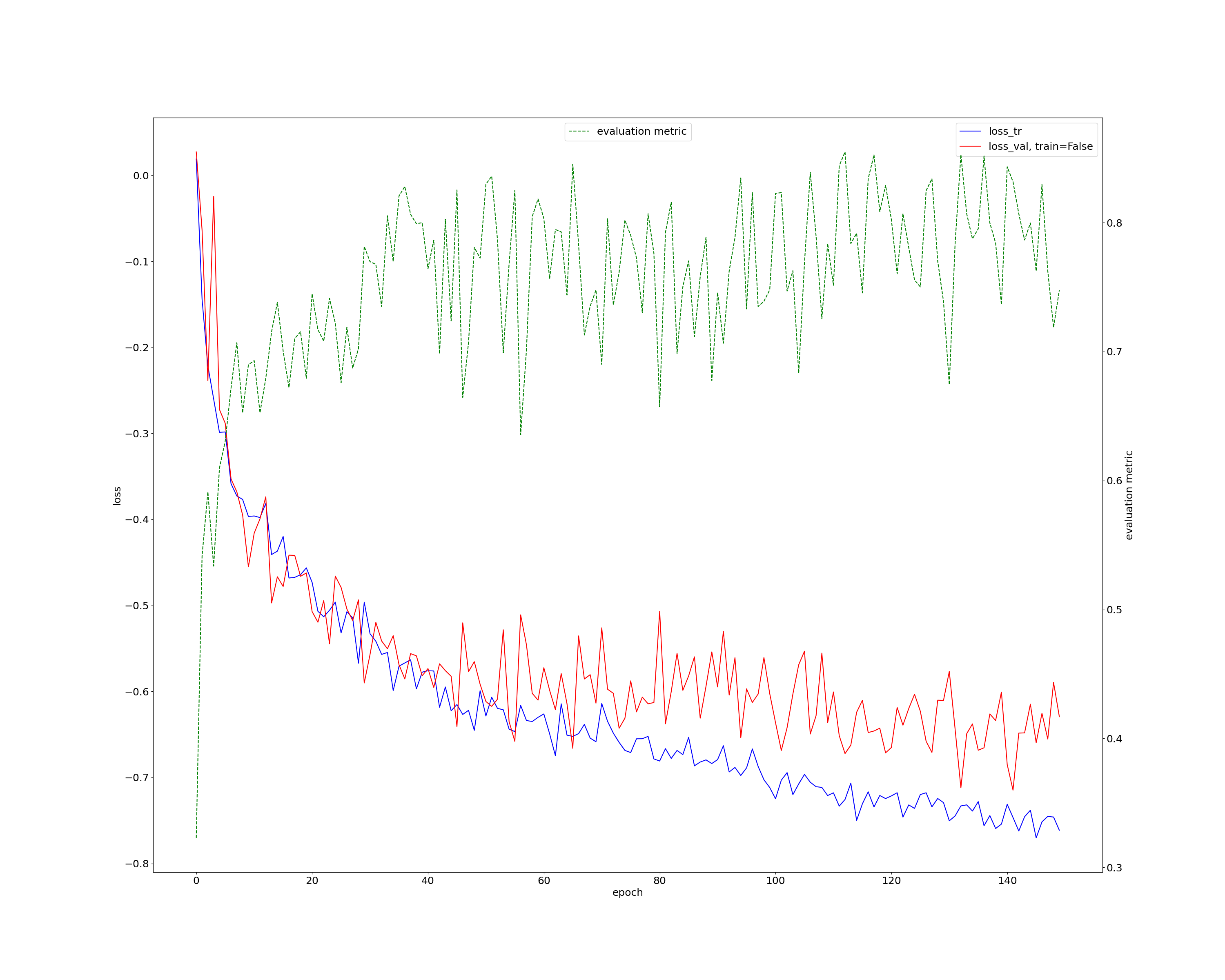}
    \caption*{\small VMamba — Fold 0}
  \end{subfigure}
  \hfill
  \begin{subfigure}{0.29\linewidth}
    \centering
    \includegraphics[width=\linewidth]{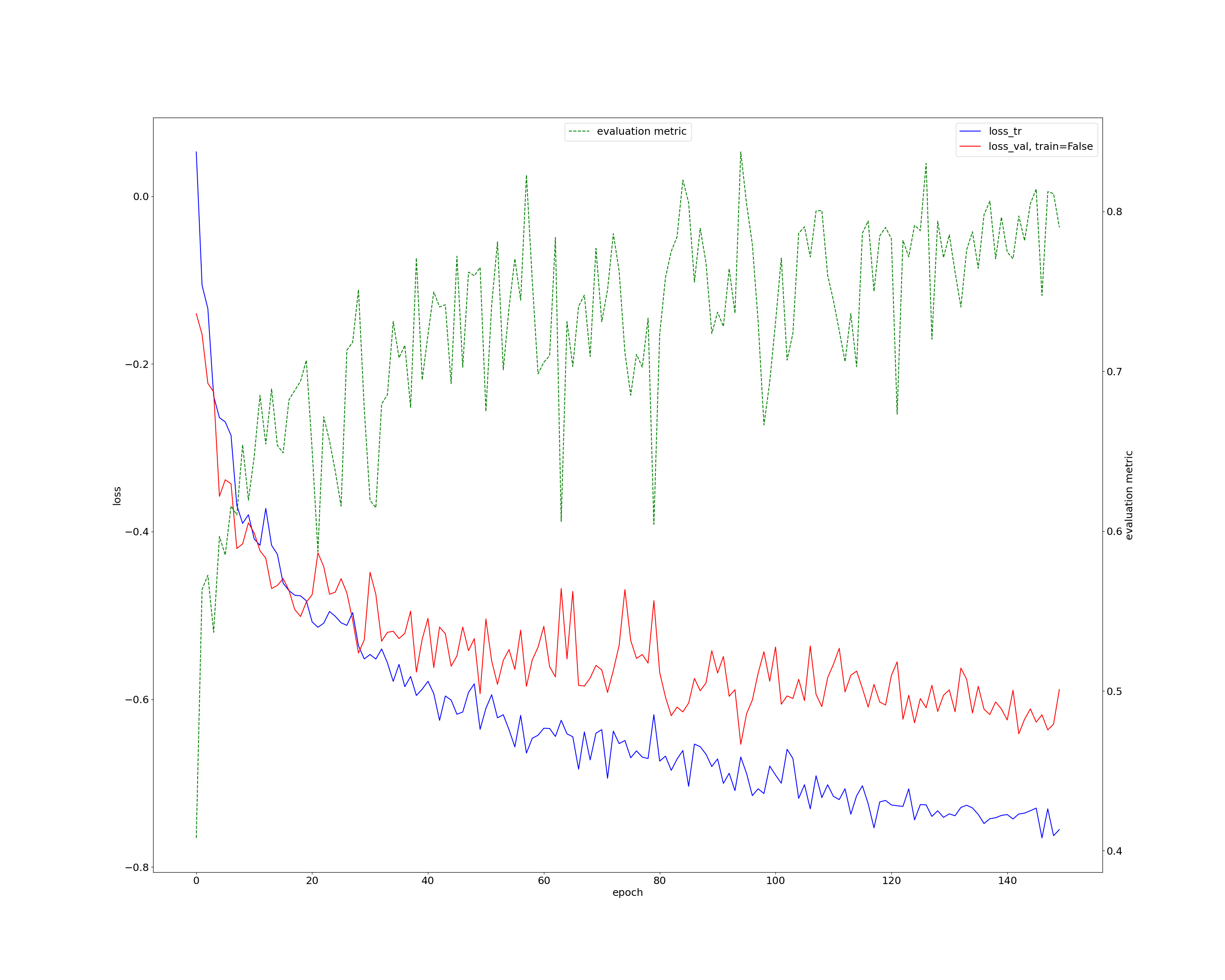}
    \caption*{\small VMamba — Fold 1}
  \end{subfigure}
  \hfill
  \begin{subfigure}{0.29\linewidth}
    \centering
    \includegraphics[width=\linewidth]{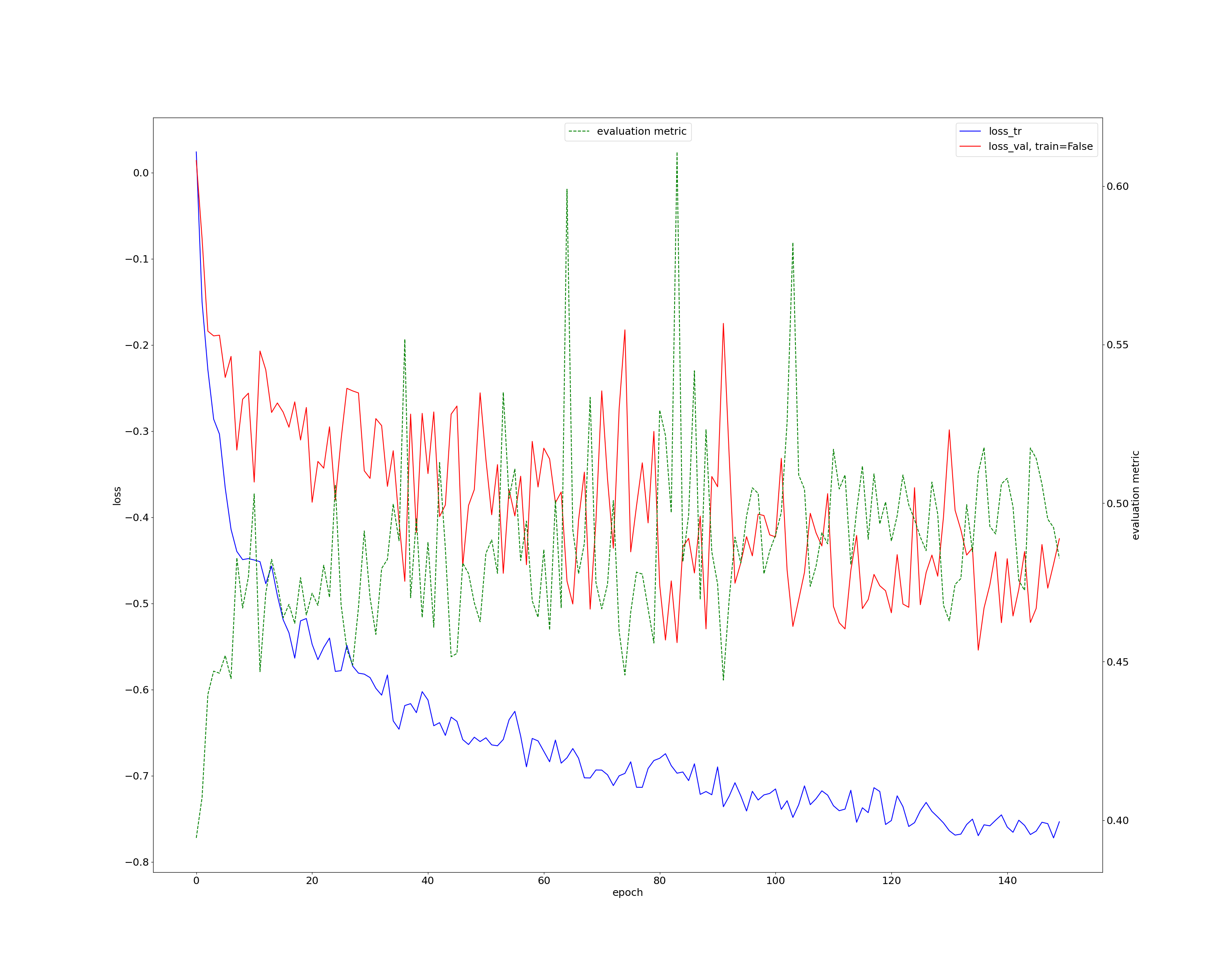}
    \caption*{\small VMamba — Fold 2}
  \end{subfigure}
  \\[3pt]

  \begin{subfigure}{0.29\linewidth}
    \centering
    \includegraphics[width=\linewidth]{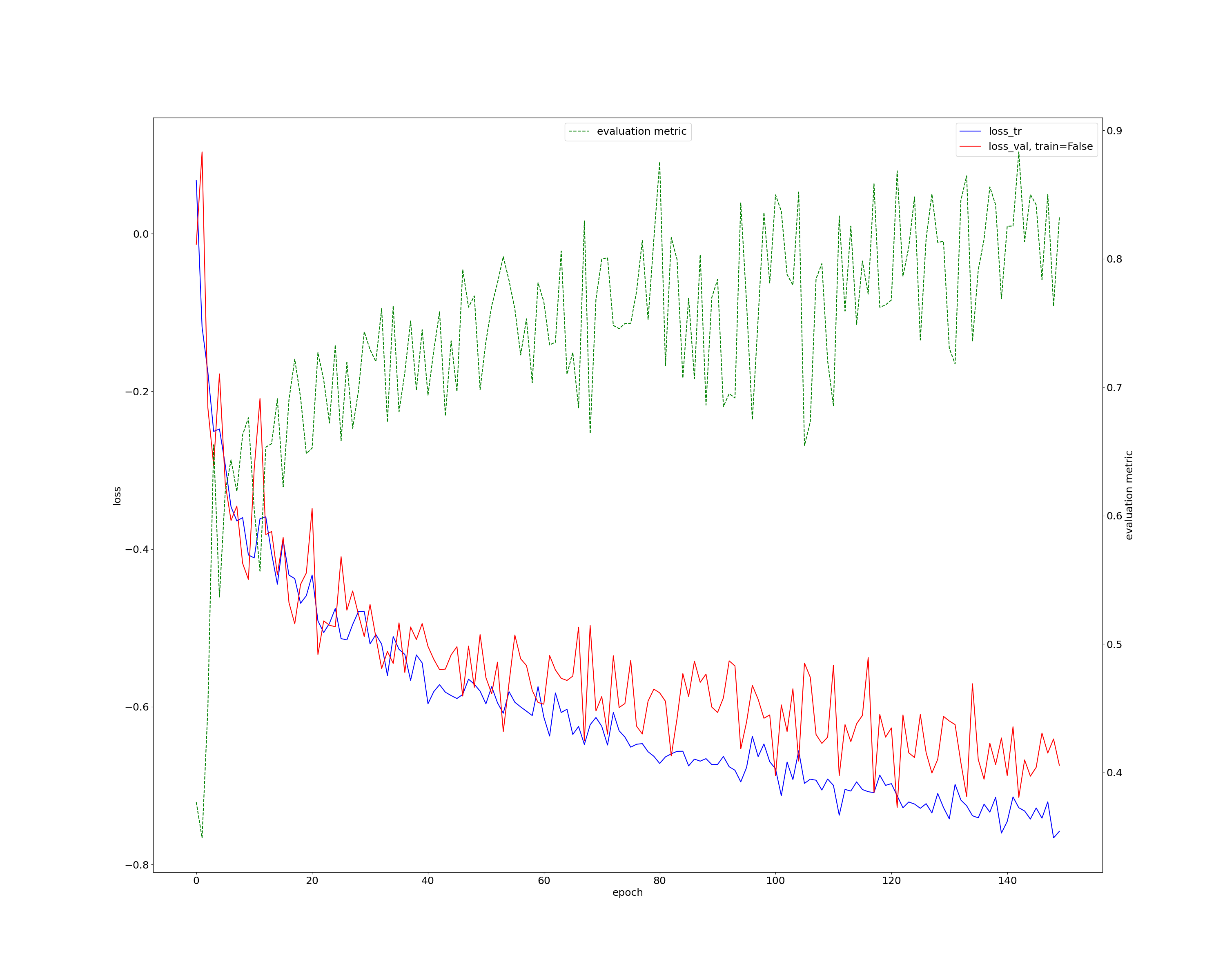}
    \caption*{\small SG + VMamba — Fold 0}
  \end{subfigure}
  \hfill
  \begin{subfigure}{0.29\linewidth}
    \centering
    \includegraphics[width=\linewidth]{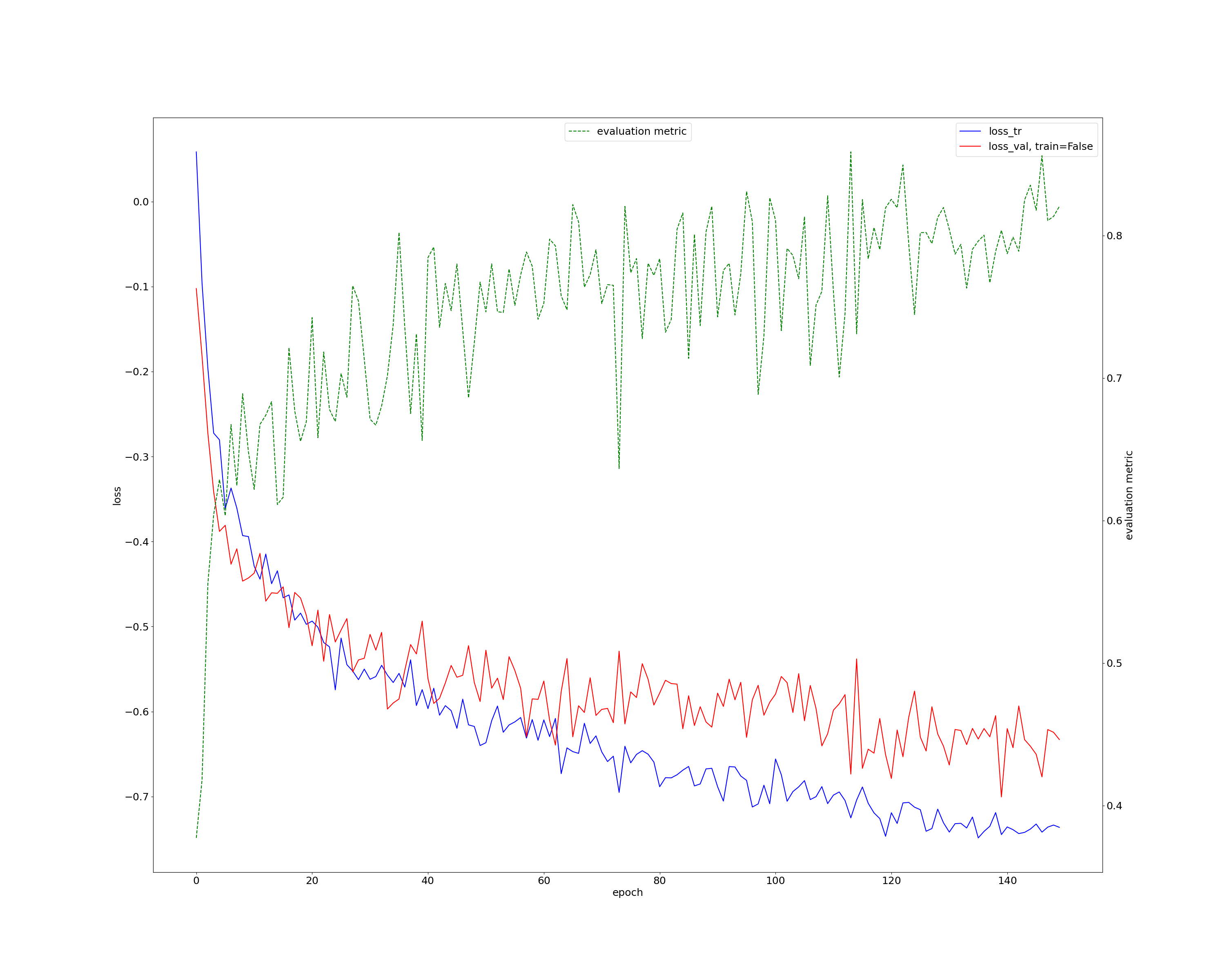}
    \caption*{\small SG + VMamba — Fold 1}
  \end{subfigure}
  \hfill
  \begin{subfigure}{0.29\linewidth}
    \centering
    \includegraphics[width=\linewidth]{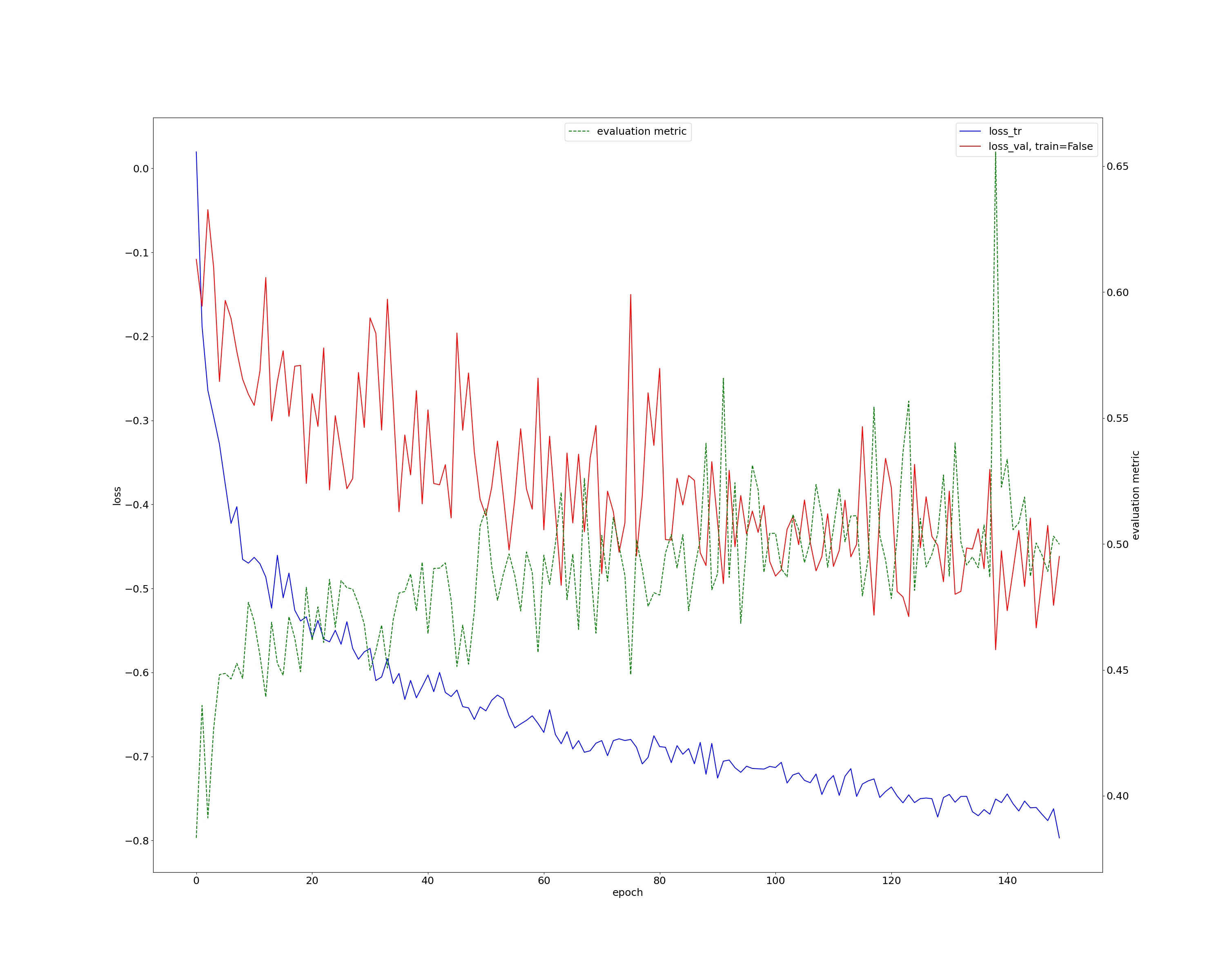}
    \caption*{\small SG + VMamba — Fold 2}
  \end{subfigure}

  \caption{LiTs Loss for 150 epochs across 3 Folds.}
  \label{fig:lits_loss}
\end{figure*}

\textbf{LiTS Convergence Challenges.} In contrast to PROMISE, Figure~\ref{fig:lits_loss} reveals that LiTS curves display higher-frequency oscillations and occasional loss spikes. These irregularities reflect the dataset's greater complexity—irregular tumor morphology, higher inter-slice variability, and our reduced sample size (50 cases versus 131 in the full dataset). The non-monotonic oscillations are characteristic of underfit or noisy gradient updates, suggesting that longer training schedules (500 vs 150 epochs) or refined learning-rate schedules may be needed for complete convergence on this challenging dataset.

\section{Fusion Gating Technical Details}\label{app:B}

\begin{enumerate}[topsep=0pt, itemsep=0.10pt, leftmargin=*]
    \item \textit{Channel gate:} A global confidence score is estimated for each channel, independent of spatial location. We first compute global average pooled (GAP) descriptors:
\footnotesize
\begin{multline*}
    \bar{x}^{2D}=\text{GAP}(x^{2D}), \quad \bar{x}^{3D}=\text{GAP}(x^{3D}), \\ \bar{x}^{2D},\bar{x}^{3D} \in \mathbb{R}^{N\times C\times 1\times 1\times 1}
\end{multline*}
\normalsize
The two descriptors are concatenated and passed through a two-layer bottleneck MLP:
\begin{equation*}
    g_c = \sigma\left(W_2 \phi \left( W_1 [\bar{x}^{2D},\bar{x}^{3D} ] \right) \right)
\end{equation*}
where $W_1, W_2$ are $1\times 1\times 1$ convolutional layers, $\phi(\cdot)$ denotes a ReLU nonlinearity, and $\sigma(\cdot)$ is a sigmoid activation. The resulting $g_c \in [0,1]^{N\times C\times D\times H\times W}$ is broadcast spatially:
\begin{equation*}
    g = g_c \otimes\textbf{1}_{D,H,W}
\end{equation*}
  \item \textit{Spatial gate:} Here the gate is computed for each spatial location, shared across channels. We first compute channel-wise mean and max feature maps:
  \begin{equation*}
      \mbox{\footnotesize $\displaystyle
  \begin{aligned}
      s_{avg} &= \text{mean}_c \left(\sfrac{1}{2}\left(x^{2D}+x^{3D}\right)\right) \\ s_{max} &= \text{max}_c \left(\sfrac{1}{2}\left(x^{2D}+x^{3D}\right)\right)
  \end{aligned}
    $}
  \end{equation*}
followed by a $1\times 1\times 1$ convolution and sigmoid activation:
\begin{equation*}
    g_s = \sigma\left(W_1 \left( [s_{avg},s_{max}] \right) \right)
\end{equation*}
where $[\cdot,\cdot]$ denotes channel concatenation. The resulting  $g_s \in [0,1]^{N\times C\times D\times H\times W}$ is broadcast overall all channels:
\begin{equation*}
    g = g_s \otimes\textbf{1}_{C}
\end{equation*}
\end{enumerate}
The gates are initialized neutrally ($g \approx 0.5$) to emulate the baseline \textit{subtraction} fusion behaviour at initialization, ensuring smooth optimization.

\end{document}